\documentclass[letterpaper]{article}
\usepackage[latin9]{inputenc}
\usepackage{float}
\usepackage{url}
\usepackage{amsmath}
\usepackage{amsthm}
\usepackage{amssymb}
\usepackage{graphicx}
\usepackage{esint}

\makeatletter

\pdfpageheight\paperheight
\pdfpagewidth\paperwidth

\floatstyle{ruled}
\newfloat{algorithm}{tbp}{loa}
\providecommand{\algorithmname}{Algorithm}
\floatname{algorithm}{\protect\algorithmname}

\theoremstyle{plain}
\newtheorem{thm}{\protect\theoremname}
\theoremstyle{plain}
\newtheorem{prop}[thm]{\protect\propositionname}
\ifx\proof\undefined
\newenvironment{proof}[1][\protect\proofname]{\par
\normalfont\topsep6\p@\@plus6\p@\relax
\trivlist
\itemindent\parindent
\item[\hskip\labelsep\scshape #1]\ignorespaces
}{%
\endtrivlist\@endpefalse
}
\providecommand{\proofname}{Proof}
\fi

\usepackage{nips15submit_e,times}
\usepackage{hyperref}
\usepackage{url}
\usepackage[square,sort,comma,numbers]{natbib}
\title{Kamiltonian Monte Carlo}

\author{
Heiko Strathmann$^*$ Dino Sejdinovic$^+$ Samuel Livingstone$^o$  Zoltan Szabo$^*$ Arthur Gretton$^*$\\
\And
\\
$^*$Gatsby Unit\\
University College London \\
\And
\\
$^+$Department of Statistics\\
University of Oxford \\
\And
\\
$^o$School of Mathematics\\
University of Bristol
}

\nipsfinalcopy % Uncomment for camera-ready version

\makeatother

\providecommand{\propositionname}{Proposition}
\providecommand{\theoremname}{Theorem}

\begin{document}
\global\long\def\diag{\text{diag}}

\title{Gradient-free Hamiltonian Monte Carlo\\
with Efficient Kernel Exponential Families}
\maketitle
\begin{abstract}
\vspace{-.3cm}We propose \emph{Kernel Hamiltonian Monte Carlo (KMC)},
a gradient-free adaptive MCMC algorithm based on Hamiltonian Monte
Carlo (HMC). On target densities where classical HMC is not an option
due to intractable gradients, KMC adaptively learns the target's gradient
structure by fitting an exponential family model in a Reproducing
Kernel Hilbert Space. Computational costs are reduced by two novel
efficient approximations to this gradient. While being asymptotically
exact, KMC mimics HMC in terms of sampling efficiency, and offers
substantial mixing improvements over state-of-the-art gradient free
samplers. We support our claims with experimental studies on both
toy and real-world applications, including Approximate Bayesian Computation
and exact-approximate MCMC.\vspace{-.4cm}
\end{abstract}

\section{Introduction}

\vspace{-.4cm}Estimating expectations using Markov Chain Monte Carlo
(MCMC) is a fundamental approximate inference technique in Bayesian
statistics.  MCMC itself can be computationally demanding, and the
expected estimation error depends directly on the correlation between
successive points in the Markov chain. Therefore, efficiency can be
achieved by taking large steps with high probability.

Hamiltonian Monte Carlo \cite{neal2011mcmc} is an MCMC algorithm
that improves efficiency by exploiting gradient information. It simulates
particle movement along the contour lines of a dynamical system constructed
from the target density. Projections of these trajectories cover wide
parts of the target's support, and the probability of accepting a
move along a trajectory is often close to one. Remarkably, this property
is mostly invariant to growing dimensionality, and HMC here often
is superior to random walk methods, which need to decrease their step
size at a much faster rate \cite[Sec. 4.4]{neal2011mcmc}.

Unfortunately, for a large class of problems, gradient information
is not available. For example, in Pseudo-Marginal MCMC (PM-MCMC) \cite{beaumont2003estimation,Andrieu2009a},
the posterior does not have an analytic expression, but can only be
estimated at any given point, e.g. in Bayesian Gaussian Process classification
\cite{FilipponeIEEETPAMI13}. A related setting is MCMC for Approximate
Bayesian Computation (ABC-MCMC), where the posterior is approximated
through repeated simulation from a likelihood model \cite{marjoram2003markov,Sisson2010}.
In both cases, HMC cannot be applied, leaving random walk methods
as the only mature alternative. There have been efforts to mimic HMC's
behaviour using stochastic gradients from mini-batches in Big Data
\cite{ChenFoxGuestrin2014}, or stochastic finite differences in ABC
\cite{Meeds2015}. Stochastic gradient based HMC methods, however,
often suffer from low acceptance rates or additional bias that is
hard to quantify \cite{Betancourt2015a}.

Random walk methods can be tuned by matching scaling of steps and
target. For example, Adaptive Metropolis-Hastings (AMH)\emph{ }\cite{Haario1999,Andrieu2008}
is based on learning the global scaling of the target from the history
of the Markov chain.  Yet, for densities with nonlinear support,
this approach does not work very well. Recently, \cite{sejdinovic_kernel_2014}
introduced a Kernel Adaptive Metropolis-Hastings (KAMH) algorithm
whose proposals are locally aligned to the target. By adaptively learning
target covariance in a Reproducing Kernel Hilbert Space (RKHS), KAMH
achieves improved sampling efficiency.

In this paper, we extend the idea of using kernel methods to learn
efficient proposal distributions \cite{sejdinovic_kernel_2014}. Rather
than \emph{locally} smoothing the target density, however, we estimate
its gradients \emph{globally}. More precisely, we fit an infinite
dimensional exponential family model in an RKHS via score matching
\cite{SriFukKumGreHyv14,Hyvarinen-05}. This is a non-parametric method
of modelling the log unnormalised target density as an RKHS function,
and has been shown to approximate a rich class of density functions
arbitrarily well. More importantly, the method has been empirically
observed to be relatively robust to increasing dimensionality -- in
sharp contrast to classical kernel density estimation \cite[Sec. 6.5]{wasserman2006all}.
Gaussian Processes (GP) were also used in \cite{Rasmussen2003} as
an emulator of the target density in order to speed up HMC, however,
this requires access to the target in closed form, to provide training
points for the GP.

We require our adaptive KMC algorithm to be computationally efficient,
as it deals with high-dimensional MCMC chains of growing length. We
develop two novel approximations to the infinite dimensional exponential
family model. The first approximation, \emph{score matching lite},
is based on computing the solution in terms of a lower dimensional,
yet growing, subspace in the RKHS. KMC with score matching lite (\emph{KMC
lite}) is geometrically ergodic on the same class of targets as standard
random walks. The second approximation uses a finite dimensional feature
space (\emph{KMC finite}), combined with random Fourier features \cite{Rahimi2007}.
KMC finite is an efficient online estimator that allows to use \emph{all}
of the Markov chain history, at the cost of decreased efficiency in
unexplored regions. A choice between KMC lite and KMC finite ultimately
depends on the ability to initialise the sampler within high-density
regions of the target; alternatively, the two approaches could be
combined.

Experiments show that KMC inherits the efficiency of HMC, and therefore
mixes significantly better than state-of-the-art gradient-free adaptive
samplers on a number of target densities, including on synthetic examples,
and when used in PM-MCMC and ABC-MCMC. All code can be found at{\footnotesize{}
\url{https://github.com/karlnapf/kernel_hmc}}\vspace{-.3cm}

\section{Background and Previous Work}

\label{sec:background_previous_work}

\vspace{-.3cm}Let the domain of interest $\mathcal{X}$ be a compact\footnote{The compactness restriction is imposed to satisfy the assumptions
in \cite{SriFukKumGreHyv14}.} subset of $\mathbb{R}^{d}$, and denote the unnormalised \emph{target}
density on $\mathcal{X}$ by $\pi$. We are interested in constructing
a Markov chain $x_{1}\to x_{2}\to\dots$ such that $\lim_{t\to\infty}x_{t}\sim\pi$.
By running the Markov chain for a long time $T$, we can consistently
approximate any expectation w.r.t. $\pi$. Markov chains are constructed
using the Metropolis-Hastings algorithm, which at the current state
$x_{t}$ draws a point from a proposal mechanism $x^{*}\sim Q(\cdot|x_{t}),$
and sets $x_{t+1}\leftarrow x^{*}$ with probability $\min(1,[\pi(x^{*})Q(x_{t}|x^{*})]/[\pi(x_{t})Q(x^{*}|x_{t})])$,
and $x_{t+1}\leftarrow x_{t}$ otherwise. We assume that $\pi$ is
intractable,\footnote{$\pi$ is analytically intractable, as opposed to computationally
expensive in the Big Data context.} i.e. that we can neither evaluate $\pi(x)$ nor\footnote{Throughout the paper $\nabla$ denotes the gradient operator w.r.t.
to $x$.} $\nabla\log\pi(x)$ for any $x$, but can only estimate it unbiasedly
via $\hat{\pi}(x)$. Replacing $\pi(x)$ with $\hat{\pi}(x)$ results
in PM-MCMC \cite{beaumont2003estimation,Andrieu2009a}, which asymptotically
remains exact (\emph{exact-approximate inference)}.

\paragraph{(Kernel) Adaptive Metropolis-Hastings}

\vspace{-.2cm}In the absence of $\nabla\log\pi$, the usual choice
of $Q$ is a random walk, i.e. $Q(\cdot|x_{t})={\cal N}(\cdot|x_{t},\Sigma_{t}).$
A popular choice of the scaling is $\Sigma_{t}\propto I$. When the
scale of the target density is not uniform across dimensions, or if
there are strong correlations, the AMH algorithm \cite{Haario1999,Andrieu2008}
improves mixing by adaptively learning global covariance structure
of $\pi$ from the history of the Markov chain. For cases where the
local scaling does not match the global covariance of $\pi$, i.e.
the support of the target is nonlinear, KAMH \cite{sejdinovic_kernel_2014}
improves mixing by learning the target covariance in a RKHS. KAMH
proposals are Gaussian with a covariance that matches the local covariance
of $\pi$ around the current state $x_{t}$, without requiring access
to $\nabla\log\pi$.

\paragraph{Hamiltonian Monte Carlo}

\vspace{-.2cm}Hamiltonian Monte Carlo (HMC) uses deterministic, measure-preserving
maps to generate efficient Markov transitions \cite{neal2011mcmc,Betancourt2015}.
Starting from the negative log target, referred to as the \emph{potential
energy} $U(q)=-\log\pi(q)$,\emph{ }we introduce an auxiliary \emph{momentum}
variable $p\sim\exp(-K(p))$ with $p\in{\cal X}$. The joint distribution
of $(p,q)$ is then proportional to $\exp\left(-H(p,q)\right)$, where
$H(p,q):=K(p)+U(q)$ is called the \emph{Hamiltonian}. $H(p,q)$ defines
a \emph{Hamiltonian flow}, parametrised by a trajectory length $t\in\mathbb{R}$,
which is a map\emph{ $\phi_{t}^{H}:(p,q)\mapsto(p^{*},q^{*})$} for
which $H(p^{*},q^{*})=H(p,q)$. This allows constructing $\pi$-invariant
Markov chains: for a chain at state $q=x_{t}$, repeatedly (i) re-sample
$p'\sim\exp(-K(\cdot))$, and then (ii) apply the Hamiltonian flow
for time $t$, giving $(p^{*},q^{*})=\phi_{t}^{H}(p',q$). The flow
can be generated by the \emph{Hamiltonian operator}
\begin{align}
\frac{\partial K}{\partial p}\frac{\partial}{\partial q}-\frac{\partial U}{\partial q}\frac{\partial}{\partial p}\label{eq:potential_energy_operator}
\end{align}
In practice, \eqref{eq:potential_energy_operator} is usually unavailable
and we need to resort to approximations. Here, we limit ourselves
to the leap-frog integrator; see \cite{neal2011mcmc} for details.
To correct for discretisation error, a Metropolis acceptance procedure
can be applied: starting from $(p',q)$, the end-point of the approximate
trajectory is accepted with probability $\min\left[1,\exp\left(-H(p^{*},q^{*})+H(p',q)\right)\right]$.
HMC is often able to propose distant, uncorrelated moves with a high
acceptance probability.

\paragraph{Intractable densities}

In many cases the gradient of $\log\pi(q)=-U(q)$ cannot be written
in closed form, leaving random-walk based methods as the state-of-the-art
\cite{sejdinovic_kernel_2014,Andrieu2008}. We aim to overcome random-walk
behaviour, so as to obtain significantly more efficient sampling \cite{neal2011mcmc}.\vspace{-.3cm}

\section{Kernel Induced Hamiltonian Dynamics}

\label{sec:kernel_hamiltonian_dynamics}\label{sec:kernel_hmc}\vspace{-.3cm}KMC
replaces the potential energy in \eqref{eq:potential_energy_operator}
by a kernel induced surrogate computed from the history of the Markov
chain. This surrogate does not require gradients of the log-target
density. The surrogate induces a kernel Hamiltonian flow, which can
be numerically simulated using standard leap-frog integration. As
with the discretisation error in HMC, any deviation of the kernel
induced flow from the true flow is corrected via a Metropolis acceptance
procedure. This here also contains the estimation noise from $\hat{\pi}$
and re-uses previous values of $\hat{\pi}$, c.f. \cite[Table 1]{Andrieu2009a}.
Consequently, the stationary distribution of the chain remains correct,
given that we take care when adapting the surrogate. \vspace{-.2cm}

\paragraph{Infinite Dimensional Exponential Families in a RKHS}

We construct a kernel induced potential energy surrogate whose gradients
approximate the gradients of the true potential energy $U$ in \eqref{eq:potential_energy_operator},
without accessing $\pi$ or $\nabla\pi$ directly, but only using
the history of the Markov chain. To that end, we model the (unnormalised)
target density $\pi(x)$ with an infinite dimensional exponential
family model \cite{SriFukKumGreHyv14} of the form
\begin{equation}
\text{const}\times\pi(x)\approx\exp\left(\langle f,k(x,\cdot)\rangle_{{\cal H}}-A(f)\right),\label{eq:infinite_exp_family}
\end{equation}

which in particular implies $\nabla f\approx-\nabla U=\nabla\log\pi.$
Here ${\cal H}$ is a RKHS of real valued functions on ${\cal X}$.
The RKHS has a uniquely associated symmetric, positive definite \emph{kernel}
$k:{\cal X}\times{\cal X}\rightarrow\mathbb{R}$, which satisfies
$f(x)=\langle f,k(x,\cdot)\rangle_{{\cal H}}$ for any $f\in{\cal H}$
\cite{BerTho04}. The canonical feature map $k(\cdot,x)\in{\cal H}$
here takes the role of the \emph{sufficient statistics} while $f\in{\cal H}$
are the \emph{natural parameters}, and $A(f):=\log\int_{{\cal X}}\exp(\langle f,k(x,\cdot)\rangle_{{\cal H}})dx$
is the cumulant generating function. Eq. \eqref{eq:infinite_exp_family}
defines broad class of densities: when universal kernels are used,
the family is dense in the space of continuous densities on compact
domains, with respect to e.g. Total Variation and KL \cite[Section 3]{SriFukKumGreHyv14}.
It is possible to consistently fit an \emph{unnormalised} version
of \eqref{eq:infinite_exp_family} by directly minimising the expected
gradient mismatch between the model \eqref{eq:infinite_exp_family}
and the true target density $\pi$ (observed through the Markov chain
history). This is achieved by generalising the score matching approach
\cite{Hyvarinen-05} to infinite dimensional parameter spaces. The
technique avoids the problem of dealing with the intractable $A(f)$,
and reduces the problem to solving a linear system. More importantly,
the approach is observed to be relatively robust to increasing dimensions.
We return to estimation in Section \ref{sec:estimators}, where we
develop two efficient approximations. For now, assume access to an
$\hat{f}\in{\cal H}$ such that $\nabla f(x)\approx\nabla\log\pi(x)$.

\paragraph{Kernel Induced Hamiltonian Flow}

\label{sub:kernel_hmc_flow}

\begin{figure}
\begin{centering}
\includegraphics[bb=50bp 0bp 175bp 144bp,clip,scale=0.55]{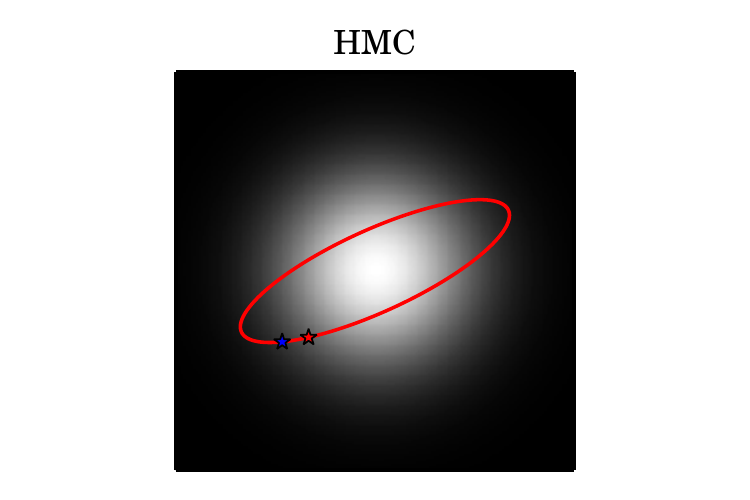}\includegraphics[scale=0.55]{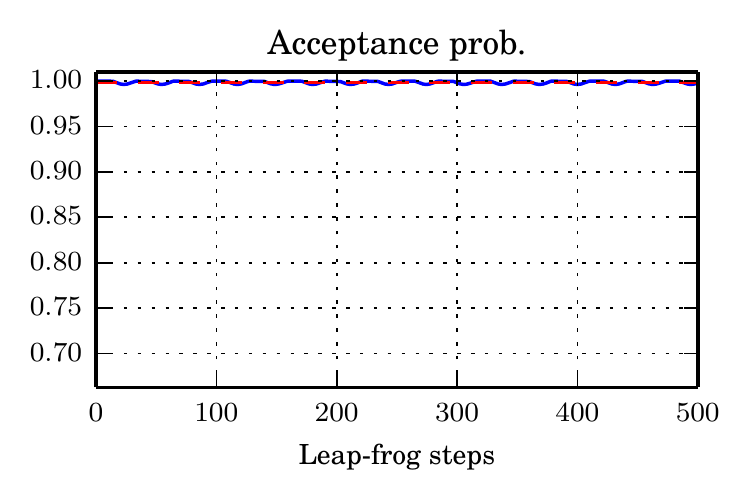}\hspace{.1cm}\includegraphics[bb=50bp 0bp 175bp 144bp,clip,scale=0.55]{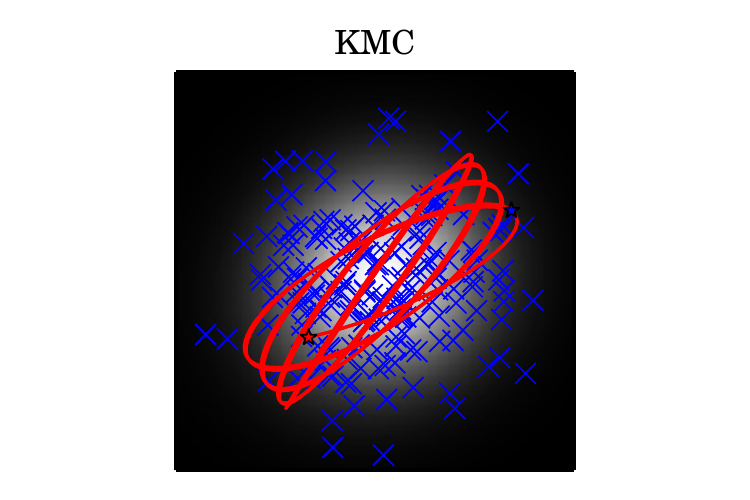}\includegraphics[scale=0.55]{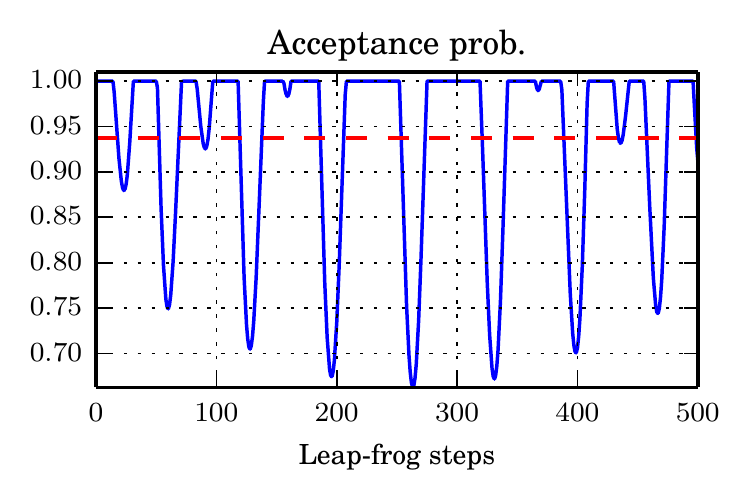}
\par\end{centering}

\caption{\label{fig:kmc_trajectories}Hamiltonian trajectories on a 2-dimensional
standard Gaussian. End points of such trajectories (red stars to blue
stars) form the proposal of HMC-like algorithms. \textbf{Left:} Plain
Hamiltonian trajectories oscillate on a stable orbit, and acceptance
probability is close to one. \textbf{Right:} Kernel induced trajectories
and acceptance probabilities on an estimated energy function.}
\end{figure}

We define a kernel induced Hamiltonian operator by replacing $U$
in the potential energy part $\frac{\partial U}{\partial p}\frac{\partial}{\partial q}$
in \eqref{eq:potential_energy_operator} by our kernel surrogate $U_{k}=f$.
It is clear that, depending on $U_{k}$, the resulting kernel induced
Hamiltonian flow differs from the original one. That said, any bias
on the resulting Markov chain, in addition to discretisation error
from the leap-frog integrator, is naturally corrected for in the Pseudo-Marginal
Metropolis step. We accept an end-point $\phi_{t}^{H_{k}}(p',q)$
of a trajectory starting at $(p',q)$ along the \emph{kernel induced
}flow with probability
\begin{equation}
\min\left[1,\exp\left(-H\left(\phi_{t}^{H_{k}}(p',q)\right)+H(p',q)\right)\right],\label{eq:kmc_accept_prob}
\end{equation}
where $H\left(\phi_{t}^{H_{k}}(p',q)\right)$ corresponds to the\emph{
true} Hamiltonian at $\phi_{t}^{H_{k}}(p',q)$. Here, in the Pseudo-Marginal
context, we replace both terms in the ratio in \eqref{eq:kmc_accept_prob}
by unbiased estimates, i.e., we replace $\pi(q)$ within $H$ with
an unbiased estimator $\hat{\pi}(q)$. Note that this also involves
`recycling' the estimates of $H$ from previous iterations to ensure
anyymptotic correctness, c.f. \cite[Table 1]{Andrieu2009a}. Any deviations
of the kernel induced flow from the true flow result in a decreased
acceptance probability \eqref{eq:kmc_accept_prob}. We therefore need
to control the approximation quality of the kernel induced potential
energy to maintain high acceptance probability in practice. See Figure
\ref{fig:kmc_trajectories} for an illustrative example.

\vspace{-.3cm}

\section{Two Efficient Estimators for Exponential Families in RKHS}

\label{sec:estimators}\vspace{-.3cm}We now address estimating the
infinite dimensional exponential family model \eqref{eq:infinite_exp_family}
from data. The original estimator in \cite{SriFukKumGreHyv14} has
a large computational cost. This is problematic in the adaptive MCMC
context, where the model has to be updated on a regular basis. We
propose two efficient approximations, each with its strengths and
weaknesses. Both are based on score matching.\vspace{-.3cm}

\subsection{Score Matching}

\label{sub:Appendix_score_matching}

\vspace{-.3cm}

Following \cite{Hyvarinen-05}, we model an unnormalised log probability
density $\log\pi(x)$ with a parametric model 
\begin{equation}
\log\tilde{\pi}_{Z}(x;f):=\log\tilde{\pi}(x;f)-\log Z(f),\label{eq:score_matching_parametric_model}
\end{equation}
where $f$ is a collection of parameters of yet unspecified dimension
(c.f. natural parameters of \eqref{eq:infinite_exp_family}), and
$Z(f)$ is an unknown normalising constant. We aim to find $\hat{f}$
from a set of $n$ samples\footnote{We assume a fixed sample set here but will use both the full chain
history $\{x_{i}\}_{i=1}^{t}$ or a sub-sample later. } ${\cal D}:=\{x_{i}\}_{i=1}^{n}\sim\pi$ such that $\pi(x)\approx\tilde{\pi}(x;\hat{f})\times\text{const}$.
From \cite[Eq. 2]{Hyvarinen-05}, the criterion being optimised is
the expected squared distance between gradients of the log density,
so-called \emph{score functions},\emph{ }
\[
J(f)=\frac{1}{2}\int_{{\cal X}}\pi(x)\left\Vert \nabla\log\tilde{\pi}(x;f)-\nabla\log\pi(x)\right\Vert _{2}^{2}dx,
\]
where we note that the normalising constants vanish from taking the
gradient $\nabla$. As shown in \cite[Theorem 1]{Hyvarinen-05}, it
is possible to compute an empirical version\emph{ without} accessing
$\pi(x)$ or $\nabla\log\pi(x)$ other than through observed samples,
\begin{equation}
\hat{J}(f)=\frac{1}{n}\sum_{x\in{\cal D}}\sum_{\ell=1}^{d}\left[\frac{\partial^{2}\log\tilde{\pi}(x;f)}{\partial x_{\ell}^{2}}+\frac{1}{2}\left(\frac{\partial\log\tilde{\pi}(x;f)}{\partial x_{\ell}}\right)^{2}\right].\label{eq:score_matching_objective_sample_version}
\end{equation}
Our approximations of the original model \eqref{eq:infinite_exp_family}
are based on minimising \eqref{eq:score_matching_objective_sample_version}
using approximate scores. \vspace{-.6cm}

\subsection{Infinite Dimensional Exponential Families Lite}

\vspace{-.2cm}The original estimator of $f$ in \eqref{eq:infinite_exp_family}
takes a dual form in a RKHS sub-space spanned by $nd+1$ kernel derivatives,
\cite[Thm. 4]{SriFukKumGreHyv14}. The update of the proposal at the
iteration $t$ of MCMC requires inversion of a $(td+1)\times(td+1)$
matrix. This is clearly prohibitive if we are to run even a moderate
number of iterations of a Markov chain. Following \cite{sejdinovic_kernel_2014},
we take a simple approach to avoid prohibitive computational costs
in $t$: we form a proposal using a random sub-sample of fixed size
$n$ from the Markov chain history, $\mathbf{z}:=\{z_{i}\}_{i=1}^{n}\subseteq\{x_{i}\}_{i=1}^{t}$.
In order to avoid excessive computation when $d$ is large, we replace
the full dual solution with a solution in terms of $\text{span}\left(\left\{ k(z_{i},\cdot)\right\} _{i=1}^{n}\right)$,
which covers the support of the true density by construction, and
grows with increasing $n$.  That is, we assume that the model \eqref{eq:score_matching_parametric_model}
takes the `light' form\vspace{-.2cm}
\begin{equation}
f(x)=\sum_{i=1}^{n}\alpha_{i}k(z_{i},x)\vspace{-.2cm},\label{eq:infinite_exp_family_lite}
\end{equation}
where $\alpha\in\mathbb{R}^{n}$ are real valued parameters that are
obtained by minimising the empirical score matching objective \eqref{eq:score_matching_objective_sample_version}.
This representation is of a form similar to \cite[Section 4.1]{Hyvarinen-07},
the main differences being that the basis functions are chosen randomly,
the basis set grows with $n$, and we will require an additional regularising
term. The estimator is summarised in the following proposition, which
is proved in Appendix \ref{sub:Appendix_lite_details}.
\begin{prop}
\label{prop:lite_estimator}Given a set of samples $\mathbf{z}=\{z_{i}\}_{i=1}^{n}$
and assuming $f(x)=\sum_{i=1}^{n}\alpha_{i}k(z_{i},x)$ for the Gaussian
kernel of the form $k(x,y)=\exp\left(-\sigma^{-1}\|x-y\|_{2}^{2}\right)$,
and $\lambda>0,$ the unique minimiser of the $\lambda\Vert f\Vert_{{\cal H}}^{2}$-regularised
empirical score matching objective \eqref{eq:score_matching_objective_sample_version}
is given by 
\begin{equation}
\hat{\alpha}_{\lambda}=-\frac{\sigma}{2}(C+\lambda I)^{-1}b,\label{eq:lite_estimator}
\end{equation}
where $b\in\mathbb{R}^{n}$ and $C\in\mathbb{R}^{n\times n}$ are
given by {\footnotesize{}
\[
b=\sum_{\ell=1}^{d}\left(\frac{2}{\sigma}(Ks_{\ell}+D_{s_{\ell}}K\mathbf{1}-2D_{x_{\ell}}Kx_{\ell})-K\mathbf{1}\right)\text{ and }C=\sum_{\ell=1}^{d}\left[D_{x_{\ell}}K-KD_{x_{\ell}}\right]\left[KD_{x_{\ell}}-D_{x_{\ell}}K\right],
\]
 }with entry-wise products $s_{\ell}:=x_{\ell}\odot x_{\ell}$ and
$D_{x}:=\text{diag}(x)$. 
\end{prop}
The estimator costs ${\cal O}(n^{3}+dn^{2})$ computation (for computing
$C,b$, and for inverting $C$) and ${\cal O}(n^{2})$ storage, for
a fixed random chain history sub-sample size $n$. This can be further
reduced via low-rank approximations to the kernel matrix and conjugate
gradient methods, which are derived in Appendix \ref{sub:Appendix_lite_details}.

Gradients of the model are given as $\nabla f(x)=\sum_{i=1}^{n}\alpha_{i}\nabla k(x,x_{i})$,
i.e. they simply require to evaluate gradients of the kernel function.
Evaluation and storage of $\nabla f(\cdot)$ both cost ${\cal O}(dn)$.\vspace{-.2cm}

\subsection{Exponential Families in Finite Feature Spaces}

\vspace{-.2cm}Instead of fitting an infinite-dimensional model on
a subset of the available data, the second estimator is based on fitting
a finite dimensional approximation using \emph{all} available data
$\{x_{i}\}_{i=1}^{t}$, in \emph{primal} form. As we will see, updating
the estimator when a new data point arrives can be done online.

Define an $m$-dimensional approximate feature space ${\cal H}_{m}=\mathbb{R}^{m}$,
and denote by $\phi_{x}\in\mathbb{{\cal H}}_{m}$ the embedding of
a point $x\in{\cal X}=\mathbb{R}^{d}$ into ${\cal H}_{m}=\mathbb{R}^{m}$.
Assume that the embedding approximates the kernel function as a finite
rank expansion $k(x,y)\approx\phi_{x}^{\top}\phi_{y}$. The log unnormalised
density of the infinite model \eqref{eq:infinite_exp_family} can
be approximated by assuming the model in \eqref{eq:score_matching_parametric_model}
takes the form
\begin{align}
f(x) & =\langle\theta,\phi_{x}\rangle_{{\cal H}_{m}}=\theta^{\top}\phi_{x}\label{eq:infinite_exp_family_random_feats}
\end{align}
To fit $\theta\in\mathbb{R}^{m}$, we again minimise the score matching
objective \eqref{eq:score_matching_objective_sample_version}, as
proved in Appendix \ref{sub:Appendix_finite_details}.
\begin{prop}
\label{prop:finite_estimator}Given a set of samples $\mathbf{x}=\{x_{i}\}_{i=1}^{t}$
and assuming $f(x)=\theta^{\top}\phi_{x}$ for a finite dimensional
feature embedding $x\mapsto\phi_{x}\in\mathbb{R}^{m}$, and $\lambda>0,$
the unique minimiser of the $\lambda\Vert\theta\Vert_{2}^{2}$-regularised
empirical score matching objective \eqref{eq:score_matching_objective_sample_version}
is given by 
\begin{equation}
\hat{\theta}_{\lambda}:=(C+\lambda I)^{-1}b,\label{eq:finite_estimator}
\end{equation}
where 
\[
b:=-\frac{1}{n}\sum_{i=1}^{t}\sum_{\ell=1}^{d}\ddot{\phi}_{x_{i}}^{\ell}\in\mathbb{R}^{m},\quad\text{}\quad C:=\frac{1}{n}\sum_{i=1}^{t}\sum_{\ell=1}^{d}\dot{\phi}_{x_{i}}^{\ell}\left(\dot{\phi}_{x_{i}}^{\ell}\right)^{T}\in\mathbb{R}^{m\times m},
\]
 with $\dot{\phi}_{x}^{\ell}:=\frac{\partial}{\partial x_{\ell}}\phi_{x}$
and $\ddot{\phi}_{x}^{\ell}:=\frac{\partial^{2}}{\partial x_{\ell}^{2}}\phi_{x}$.
\end{prop}
An example feature embedding based on random Fourier features \cite{Rahimi2007,sriperumbudurszabo15optimal}
and a standard Gaussian kernel is {\footnotesize{}$\phi_{x}=\sqrt{\frac{2}{m}}\left[\cos(\omega_{1}^{T}x+u_{1}),\dots,\cos(\omega_{m}^{T}x+u_{m})\right]$},
with {\footnotesize{}$\omega_{i}\sim{\cal N}(\omega)$} and {\footnotesize{}$u_{i}\sim\texttt{Uniform}[0,2\pi]$}.
The estimator has a one-off cost of ${\cal O}(tdm^{2}+m^{3})$ computation
and ${\cal O}(m^{2})$ storage. Given that we have computed a solution
based on the Markov chain history $\{x_{i}\}_{i=1}^{t}$, however,
it is straightforward to update $C,b$, and the solution $\hat{\theta}_{\lambda}$
online, after a new point $x_{t+1}$ arrives. This is achieved by
storing running averages and performing low-rank updates of matrix
inversions, and costs ${\cal O}(dm^{2})$ computation and ${\cal O}(m^{2})$
storage, \emph{independent} of $t$. Further details are given in
Appendix \ref{sub:Appendix_finite_details}.

Gradients of the model are $\nabla f(x)=\left[\nabla\phi_{x}\right]^{\top}\hat{\theta}$
, i.e., they require the evaluation of the gradient of the feature
space embedding, costing ${\cal O}(md)$ computation and and ${\cal O}(m)$
storage.\vspace{-.3cm}

\section{Kernel Hamiltonian Monte Carlo}

\label{sec:KMC}

\vspace{-.3cm}Constructing a kernel induced Hamiltonian flow as in
Section \ref{sub:kernel_hmc_flow} from the gradients of the infinite
dimensional exponential family model \eqref{eq:infinite_exp_family},
and approximate estimators \eqref{eq:infinite_exp_family_lite},\eqref{eq:infinite_exp_family_random_feats},
we arrive at a gradient free, adaptive MCMC algorithm: \emph{Kernel
Hamiltonian Monte Carlo} (Algorithm \ref{alg:Kamiltonian-Monte-Carlo}).
\vspace{-.2cm}

\paragraph{Computational Efficiency, Geometric Ergodicity, and Burn-in}

KMC finite using \eqref{eq:infinite_exp_family_random_feats} allows
for online updates using the \emph{full} Markov chain history, and
therefore is a more elegant solution than KMC lite, which has greater
computational cost and requires sub-sampling the chain history. Due
to the parametric nature of KMC finite, however, the tails of the
estimator are not guaranteed to decay. For example, the random Fourier
feature embedding described below Proposition \ref{prop:finite_estimator}
contains periodic cosine functions, and therefore oscillates in the
tails of \eqref{eq:infinite_exp_family_random_feats}, resulting in
a reduced acceptance probability. As we will demonstrate in the experiments,
this problem does not appear when KMC finite is initialised in high-density
regions, nor after burn-in. In situations where information about
the target density support is unknown, and during burn-in, we suggest
to use the lite estimator \eqref{eq:lite_estimator}, whose gradients
decay outside of the training data. As a result, KMC lite is guaranteed
to fall back to a Random Walk Metropolis in unexplored regions, inheriting
its convergence properties, and smoothly transitions to HMC-like proposals
as the MCMC chain grows. A proof of the proposition below can be found
in Appendix \ref{sub:Appenidx_ergodicity_lite_proof}.
\begin{prop}
\label{prop:ergodicity_kmc_lite}Assume\emph{ $d=1$,} $\pi(x)$ has
log-concave tails, the regularity conditions of \cite[Thm 2.2]{roberts1996geometric}
(implying $\pi$-irreducibility and smallness of compact sets), that
MCMC adaptation stops after a fixed time, and a fixed number $L$
of $\epsilon$-leapfrog steps. If $\limsup_{\|x\|_{2}\to\infty}\|\nabla f(x)\|_{2}=0$,
and $\exists M:\forall x:\|\nabla f(x)\|_{2}\leq M$, then KMC lite
is geometrically ergodic from $\pi$-almost any starting point.

\vspace{-1cm}
\end{prop}

\paragraph{}

\paragraph{Vanishing adaptation}

\label{sub:adaptive_subsampling}

MCMC algorithms that use the history of the Markov chain for constructing
proposals might not be asymptotically correct. We follow \cite[Sec. 4.2]{sejdinovic_kernel_2014}
and the idea of `vanishing adaptation' \cite{Andrieu2008}, to avoid
such biases. Let $\left\{ a_{t}\right\} _{i=0}^{\infty}$ be a schedule
of decaying probabilities such that $\lim_{t\to\infty}a_{t}=0$ and
$\sum_{t=0}^{\infty}a_{t}=\infty$. We update the density gradient
estimate according to this schedule in Algorithm \ref{alg:Kamiltonian-Monte-Carlo}.
Intuitively, adaptation becomes less likely as the MCMC chain progresses,
but never fully stops, while sharing asymptotic convergence with adaptation
that stops at a fixed point \cite[Theorem 1]{RobertsRosenthal2007}.
Note that Proposition \ref{prop:ergodicity_kmc_lite} is a stronger
statement about the \emph{convergence rate}.\vspace{-.2cm}

\paragraph{Free Parameters}

KMC has two free parameters: the Gaussian kernel bandwidth $\sigma$,
and the regularisation parameter $\lambda$. As KMC's performance
depends on the quality of the approximate infinite dimensional exponential
family model in \eqref{eq:infinite_exp_family_lite} or \eqref{eq:infinite_exp_family_random_feats},
a principled approach is to use the score matching objective function
in \eqref{eq:score_matching_objective_sample_version} to choose $\sigma,\lambda$
pairs via cross-validation (using e.g. `hot-started' black-box optimisation).
Earlier adaptive kernel-based MCMC methods \cite{sejdinovic_kernel_2014}
did not address parameter choice.\vspace{-.3cm}

\begin{algorithm}
\textbf{\caption{\textbf{Kernel Hamiltonian Monte Carlo -- Pseudo-code\label{alg:Kamiltonian-Monte-Carlo}}}
}

\textbf{\emph{\footnotesize{}Input}}\textbf{\footnotesize{}:}{\footnotesize{}
Target (possibly noisy estimator) $\hat{\pi}$, adaptation schedule
$a_{t}$, HMC parameters,}{\footnotesize \par}

{\footnotesize{}\hspace{1cm}Size of basis $m$ or sub-sample size
$n$.}{\footnotesize \par}

{\footnotesize{}At iteration $t+1$, current state $x_{t}$, history
$\{x_{i}\}_{i=1}^{t}$}\textit{\footnotesize{}, }\textit{\emph{\footnotesize{}perform
(1-4) with probability }}\textit{\footnotesize{}$a_{t}$}{\footnotesize \par}

{\footnotesize{}}%
\begin{minipage}[t]{0.5\textwidth}%
\textbf{\footnotesize{}KMC lite:}{\footnotesize \par}
\begin{enumerate}
\item {\footnotesize{}Update sub-sample $\mathbf{z}\subseteq\{x_{i}\}_{i=1}^{t}$}{\footnotesize \par}
\item {\footnotesize{}Re-compute $C,b$ from Prop. \ref{prop:lite_estimator}}{\footnotesize \par}
\item {\footnotesize{}Solve $\hat{\alpha}_{\lambda}=-\frac{\sigma}{2}(C+\lambda I)^{-1}b$}{\footnotesize \par}
\item {\footnotesize{}$\nabla f(x)\leftarrow\sum_{i=1}^{n}\alpha_{i}\nabla k(x,z_{i})$}{\footnotesize \par}\end{enumerate}
\end{minipage}{\footnotesize{}}%
\begin{minipage}[t]{0.5\textwidth}%
\textbf{\footnotesize{}KMC finite:}{\footnotesize \par}
\begin{enumerate}
\item {\footnotesize{}Update to $C,b$ from Prop. \ref{prop:finite_estimator}}{\footnotesize \par}
\item {\footnotesize{}Perform rank-$d$ update to $C^{-1}$}{\footnotesize \par}
\item {\footnotesize{}Update $\hat{\theta}_{\lambda}=(C+\lambda I)^{-1}b$}{\footnotesize \par}
\item {\footnotesize{}$\nabla f(x)\leftarrow\left[\nabla\phi_{x}\right]^{\top}\hat{\theta}$}{\footnotesize \par}\end{enumerate}
\end{minipage}{\footnotesize \par}
\begin{enumerate}
\item [5.]\setcounter{enumi}{3}{\footnotesize{}Propose $(p',x^{*})$ with
kernel induced Hamiltonian flow, using $\nabla_{x}U=\nabla_{x}f$}{\footnotesize \par}
\item [6.]\setcounter{enumi}{3}{\footnotesize{}Perform Metropolis step
using $\hat{\pi}$: accept $x_{t+1}\leftarrow x^{*}$ w.p. \eqref{eq:kmc_accept_prob}
and reject $x_{t+1}\leftarrow x_{t}$ otherwise}\\
{\footnotesize{}If $\hat{\pi}$ is noisy and $x^{*}$ was accepted,
store above $\hat{\pi}(x^{*})$ for evaluating \eqref{eq:kmc_accept_prob}
in the next iteration}{\footnotesize \par}\end{enumerate}
\end{algorithm}

\section{Experiments}

\label{sec:Experiments}\vspace{-.3cm}We start by quantifying performance
of KMC finite on synthetic targets. We emphasise that these results
can be reproduced with the lite version.\vspace{-.3cm}

\paragraph{KMC Finite: Stability of Trajectories in High Dimensions}

\begin{figure}
\begin{centering}
\includegraphics[scale=0.5]{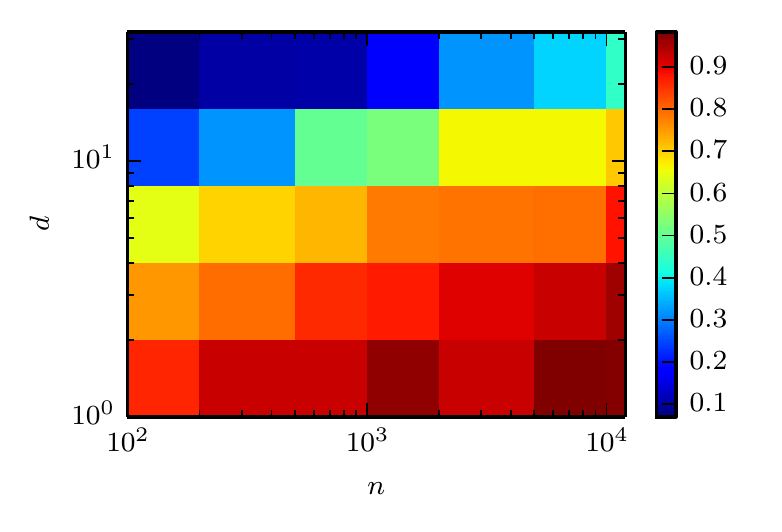}\includegraphics[scale=0.5]{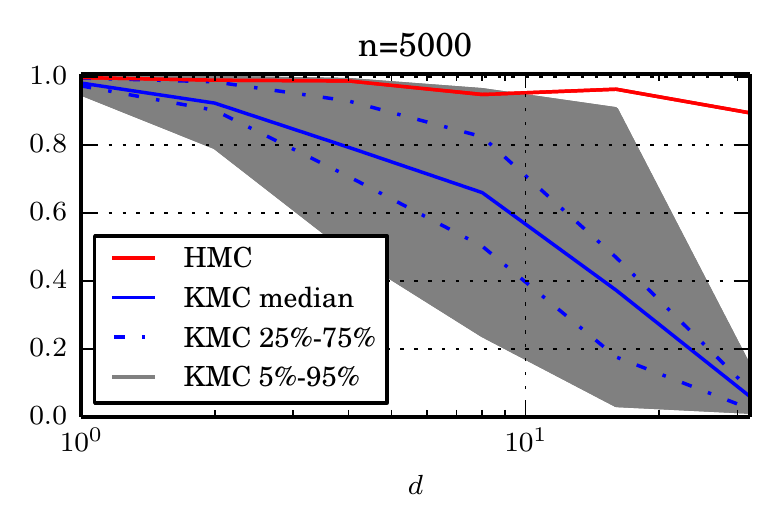}\includegraphics[bb=0bp 0bp 225bp 153bp,scale=0.5]{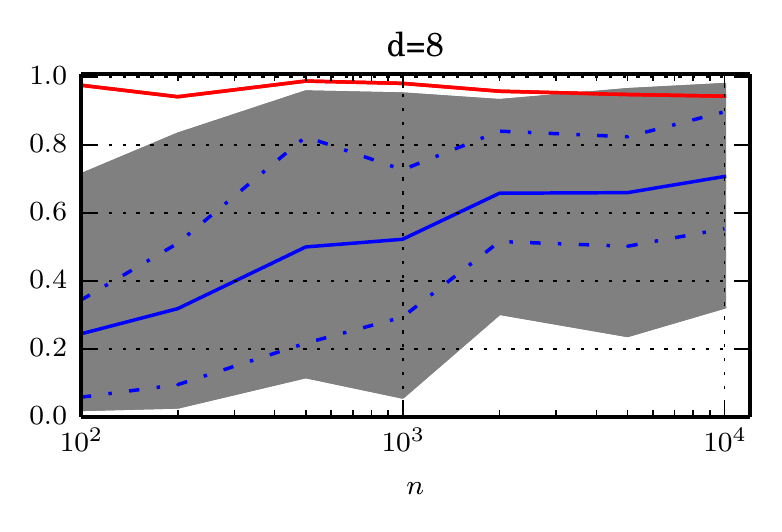}
\par\end{centering}

\caption{\label{fig:kmc_trajectories_mean_acceptance}Hypothetical acceptance
probability of KMC finite on a challening target in growing dimensions.
\textbf{Left:} As a function of\textbf{ $n=m$} (x-axis) and $d$
(y-axis). \textbf{Middle/right: }Slices through left plot with error
bars for fixed $n=m$ and as a function of \textbf{$d$} (left), and
for fixed $d$ as a function of $n=m$ (right).}
\end{figure}

In order to quantify efficiency in growing dimensions, we study hypothetical
acceptance rates along trajectories on the kernel induced Hamiltonian
flow (no MCMC yet) on a challenging Gaussian target: We sample the
diagonal entries of the covariance matrix from a \texttt{Gamma(1,1)}
distribution and rotate with a uniformly sampled random orthogonal
matrix. The resulting target is challenging to estimate due to its
`non-singular smoothness', i.e., substantially differing length-scales
across its principal components. As a single Gaussian kernel is not
able to effeciently represent such scaling families, we use a rational
quadratic kernel for the gradient estimation, whose random features
are straightforward to compute. Figure \ref{fig:kmc_trajectories_mean_acceptance}
shows the average acceptance over $100$ independent trials as a function
of the number of (ground truth) samples and basis functions, which
are set to be equal $n=m$, and of dimension $d$. In low to moderate
dimensions, gradients of the finite estimator lead to acceptance rates
comparable to plain HMC. On targets with more `regular' smoothness,
the estimator performs well in up to $d\approx100$, with less variance.
See Appendix \ref{sub:app_stability_in_high_dim} for details. \vspace{-.3cm}

\paragraph{KMC Finite: HMC-like Mixing on a Synthetic Example}

\label{sub:experiment_hmc_like_mixing}

\begin{figure}
\begin{centering}
\includegraphics[scale=0.5]{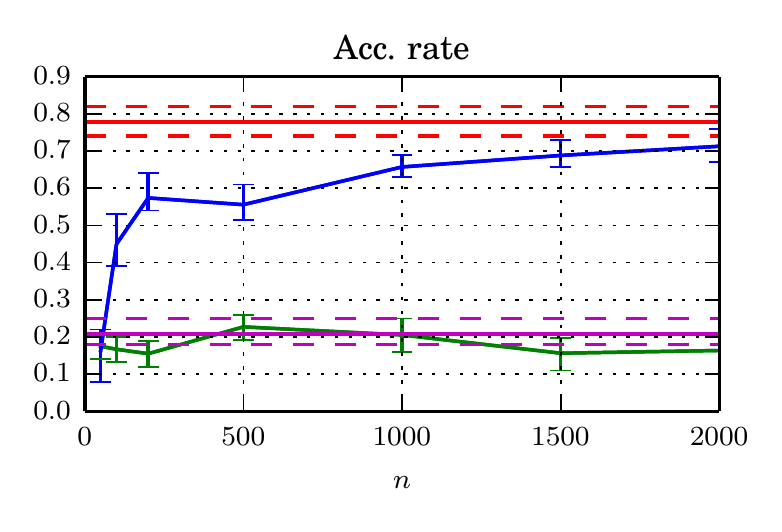}\includegraphics[scale=0.5]{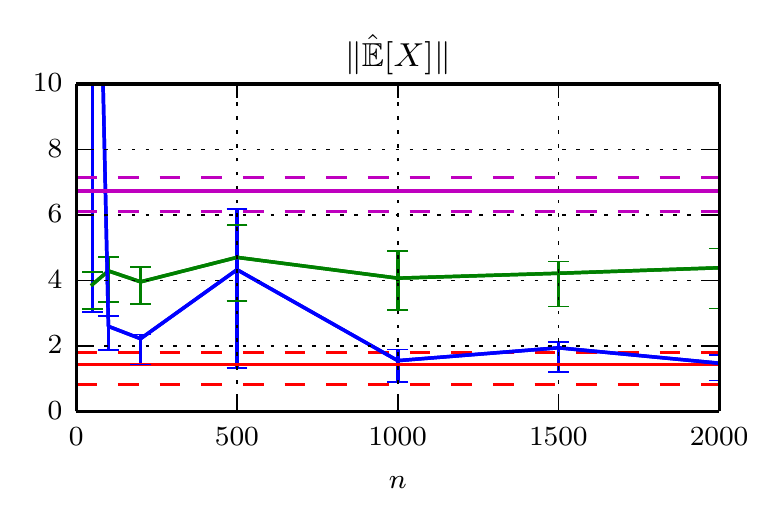}\includegraphics[scale=0.5]{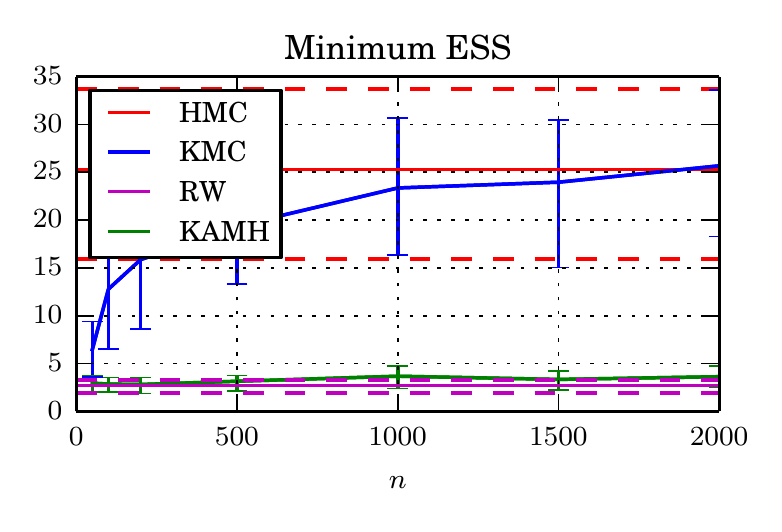}
\par\end{centering}

\caption{\label{fig:experiment_synthetic_banana}Results for the 8-dimensional
synthetic Banana. As the amout of observed data increases, KMC performance
approaches HMC -- outperforming KAMH and RW. 80\% error bars over
30 runs.}
\end{figure}

We next show that KMC's performance approaches that of HMC as it sees
more data. We compare KMC, HMC, an isotropic random walk (RW), and
KAMH on the 8-dimensional nonlinear banana-shaped target; see Appendix
\ref{sub:app_banana_target}. We here only quantify mixing \emph{after
}a sufficient burn-in (burn-in speed is included in next example).
We quantify performance on estimating the target's mean, which is
exactly $\mathbf{0}$. We tuned the scaling of KAMH and RW to achieve
23\% acceptance. We set HMC parameters to achieve 80\% acceptance
and then used the same parameters for KMC. We ran all samplers for
2000+200 iterations from a random start point, discarded the burn-in
and computed acceptance rates, the norm of the empirical mean $\mathbb{\Vert\hat{E}}[x]\Vert$,
and the minimum effective sample size (ESS) across dimensions. For
KAMH and KMC, we repeated the experiment for an increasing number
of burn-in samples and basis functions $m=n$. Figure \ref{fig:experiment_synthetic_banana}
shows the results as a function of $m=n$. KMC clearly outperforms
RW and KAMH, and eventually achieves performance close to HMC as $n=m$
grows.\vspace{-.3cm}

\paragraph{KMC Lite: Pseudo-Marginal MCMC for GP Classification on Real World
Data}

\begin{figure}
\begin{centering}
\includegraphics[bb=-5bp 0bp 225bp 153bp,scale=0.5]{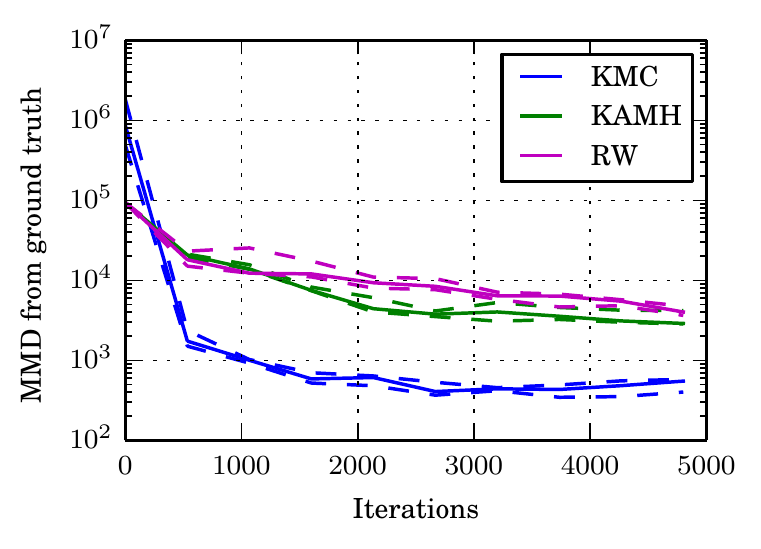}\includegraphics[scale=0.55]{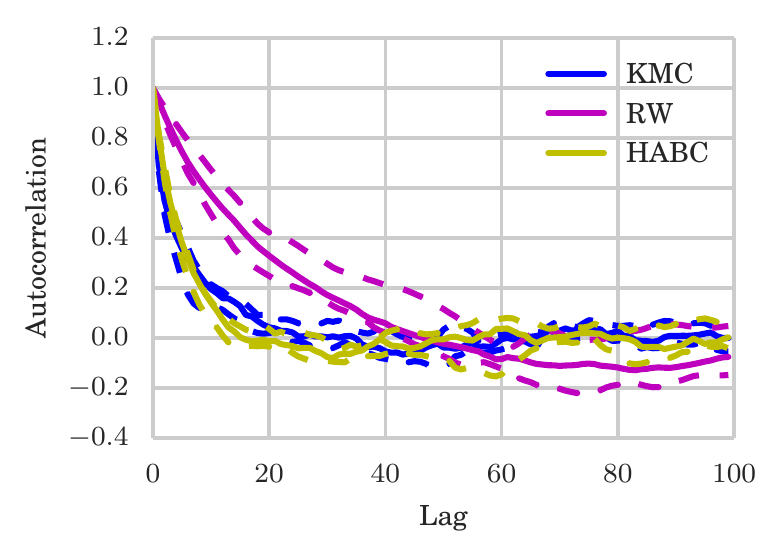}\includegraphics[scale=0.55]{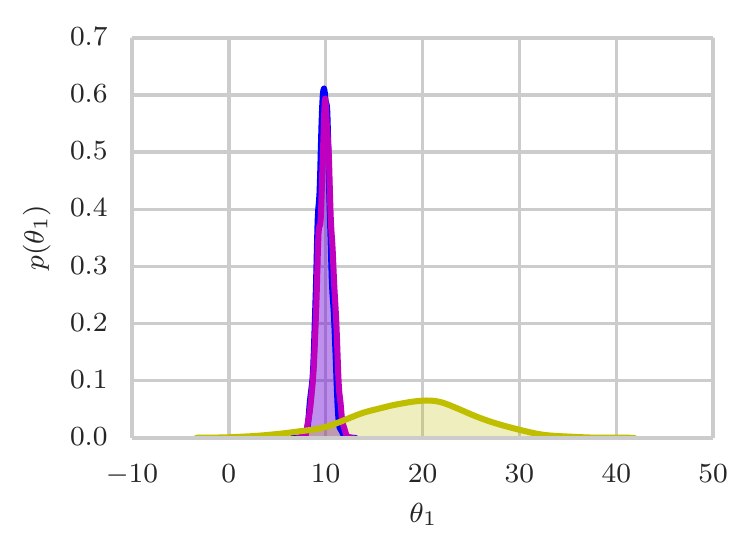}
\par\end{centering}

\caption{\label{fig:experiment_gp_abc}\textbf{Left: }Results for 9-dimensional
marginal posterior over length scales of a GPC model applied to the
UCI Glass dataset. The plots shows convergence (no burn-in discarded)
of all mixed moments up to order 3 (lower MMD is better). \textbf{Middle/right:}
ABC-MCMC auto-correlation and marginal $\theta_{1}$ posterior for
a 10-dimensional skew normal likelihood. While KMC mixes as well as
HABC, it does not suffer from any bias (overlaps with RW, while HABC
is significantly different) and requires fewer simulations per proposal.
}
\end{figure}

We next apply KMC to sample from the marginal posterior over hyper-parameters
of a Gaussian Process Classification (GPC) model on the UCI Glass
dataset \cite{Bache2013}. Classical HMC cannot be used for this problem,
due to the intractability of the marginal data likelihood. Our experimental
protocol mostly follows \cite[Section 5.1]{sejdinovic_kernel_2014},
see Appendix \ref{sub:app_gp_target}, but uses only 6000 MCMC iterations
\emph{without} discarding a burn-in, i.e., we study how fast KMC initially
explores the target. We compare convergence in terms of all mixed
moments of order up to 3 to a set of benchmark samples (MMD \cite{Grettonetal12},
lower is better). KMC randomly uses between 1 and 10 leapfrog steps
of a size chosen uniformly in $[0.01,0.1]$, a standard Gaussian momentum,
and a kernel tuned by cross-validation, see Appendix \ref{sub:app_gp_target}.
We did not extensively tune the HMC parameters of KMC as the described
settings were sufficient. Both KMC and KAMH used 1000 samples from
the chain history. Figure \ref{fig:experiment_gp_abc} (left) shows
that KMC's burn-in contains a short `exploration phase' where produced
estimates are bad, due to it falling back to a random walk in unexplored
regions, c.f. Proposition \ref{prop:ergodicity_kmc_lite}. From around
500 iterations, however, KMC clearly outperforms both RW and the earlier
state-of-the-art KAMH. These results are backed by the minimum ESS
(not plotted), which is around 415 for KMC and is around 35 and 25
for KAMH and RW, respectively. Note that all samplers effectively
stop improving from 3000 iterations --  indicating a burn-in bias.
All samplers took 1h time, with most time spent estimating the marginal
likelihood.\vspace{-.4cm}

\paragraph{KMC Lite: Reduced Simulations and no Additional Bias in ABC}

\label{sub:experiment_abc}

We now apply KMC in the context of Approximate Bayesian Computation
(ABC), which often is employed when the data likelihood is intractable
but can be obtained by simulation, see e.g. \cite{Sisson2010}. ABC-MCMC
\cite{marjoram2003markov} targets an approximate posterior by constructing
an unbiased Monte Carlo estimator of the approximate likelihood. As
each such evaluation requires expensive simulations from the likelihood,
the goal of all ABC methods is to reduce the number of such simulations.
Accordingly, Hamiltonian ABC was recently proposed \cite{Meeds2015},
combining the synthetic likelihood approach \cite{Wood:2010aa} with
gradients based on stochastic finite differences. We remark that this
requires to simulate from the likelihood in \emph{every }leapfrog
step, and that the additional bias from the Gaussian likelihood approximation
can be problematic. In contrast, KMC does not require simulations
to construct a proposal, but rather `invests' simulations into an
accept/reject step \eqref{eq:kmc_accept_prob} that ensures convergence
to the \emph{original} ABC target. Figure \ref{fig:experiment_gp_abc}
(right) compares performance of RW, HABC (sticky random numbers and
SPAS, \cite[Sec. 4.3, 4.4]{Meeds2015}), and KMC on a $10$-dimensional
skew-normal distribution $p(y|\theta)=2\mathcal{N}\left(\theta,I\right)\Phi\left(\left\langle \alpha,y\right\rangle \right)$
with $\theta=\alpha=\mathbf{1}\cdot10$. KMC mixes as well as HABC,
but HABC suffers from a severe bias. KMC also reduces the number of
simulations per proposal by a factor $2L=100$. See Appendix \ref{sub:app_abc_mcmc}
for details.\vspace{-.5cm}

\section{Discussion}

\label{sec:Discussion}\vspace{-.3cm}We have introduced KMC, a kernel-based
gradient free adaptive MCMC algorithm that mimics HMC's behaviour
by estimating target gradients in an RKHS. In experiments, KMC outperforms
random walk based sampling methods in up to $d=50$ dimensions, including
the recent kernel-based KAMH  \cite{sejdinovic_kernel_2014}. KMC
is particularly useful when gradients of the target density are unavailable,
as in PM-MCMC or ABC-MCMC, where classical HMC cannot be used. We
have proposed two efficient empirical estimators for the target gradients,
each with different strengths and weaknesses, and have given experimental
evidence for the robustness of both.

Future work includes establishing theoretical consistency and uniform
convergence rates for the empirical estimators, for example via using
recent analysis of random Fourier Features with tight bounds \cite{sriperumbudurszabo15optimal},
and a thorough experimental study in the ABC-MCMC context where we
see a lot of potential for KMC. It might also be possible to use KMC
as a precomputing strategy to speed up classical HMC as in \cite{zhang2015hamiltonian}.
For code, see{\footnotesize{} \url{https://github.com/karlnapf/kernel_hmc}}{\footnotesize \par}

\newpage{}

\bibliographystyle{unsrt}
\bibliography{local}

\normalsize\onecolumn

\part*{Appendix}

The Appendix contains proofs for Propositions \ref{prop:lite_estimator}
and \ref{prop:finite_estimator}, as well as additional computational
details for both KMC lite in Section \ref{sub:Appendix_lite_details}
and KMC finite in Section \ref{sub:Appendix_finite_details}. Section
\ref{sub:Appenidx_ergodicity_lite_proof} covers the proof of geometric
ergodicity of KMC lite from Proposition \ref{prop:ergodicity_kmc_lite}.
Section \ref{sec:app_experimental_details} describes further experimental
details.

\appendix

\section{Lite Estimator}

\label{sub:Appendix_lite_details}

\subsubsection{Proof of Proposition \ref{prop:lite_estimator}}

The proof below extends the model in \cite[Section 4.1]{Hyvarinen-07}.
We assume that the model log-density \eqref{eq:score_matching_parametric_model}
takes the form in Proposition \ref{prop:lite_estimator}, then directly
implement score functions \eqref{eq:score_matching_objective_sample_version},
from which we derive an empirical score matching objective as a system
of linear equations.
\begin{proof}
As assumed the log unnormalised density takes the form

\[
f(x)=\sum_{i=1}^{n}\alpha_{i}k(x_{i},x)
\]
where $k:\mathbb{R}^{d}\times\mathbb{R}^{d}\to\mathbb{R}$ is the
Gaussian kernel in the form
\[
k(x_{i},x)=\exp\left(-\frac{1}{\sigma}\|x_{i}-x\|^{2}\right)=\exp\left(-\frac{1}{\sigma}\sum_{\ell=1}^{d}(x_{i\ell}-x_{\ell})^{2}\right).
\]
The score functions for \eqref{eq:score_matching_objective_sample_version}
are then given by
\[
\psi_{\ell}(x;\alpha):=\frac{\partial\log\tilde{\pi}(x;f)}{\partial x_{\ell}}=\frac{2}{\sigma}\sum_{i=1}^{n}\alpha_{i}(x_{i\ell}-x_{\ell})\exp\left(-\frac{\|x_{i}-x\|^{2}}{\sigma}\right),
\]
and
\begin{align*}
\partial_{\ell}\psi_{\ell}(x;\alpha):= & \frac{\partial^{2}\log\tilde{\pi}(x;f)}{\partial^{2}x_{\ell}}\\
 & =-\frac{2}{\sigma}\sum_{i=1}^{n}\alpha_{i}\exp\left(-\frac{\|x_{i}-x\|^{2}}{\sigma}\right)+\left(\frac{2}{\sigma}\right)^{2}\sum_{i=1}^{n}\alpha_{i}(x_{i\ell}-x_{\ell})^{2}\exp\left(-\frac{\|x_{i}-x\|^{2}}{\sigma}\right)\\
 & =\frac{2}{\sigma}\sum_{i=1}^{n}\alpha_{i}\exp\left(-\frac{\|x_{i}-x\|^{2}}{\sigma}\right)\left[-1+\frac{2}{\sigma}(x_{i\ell}-x_{\ell})^{2}\right].
\end{align*}
Substituting this into \eqref{eq:score_matching_objective_sample_version}
yields
\begin{align*}
J(\alpha) & =\frac{1}{n}\sum_{i=1}^{n}\sum_{\ell=1}^{d}\left[\partial_{\ell}\psi_{\ell}(x_{i};\alpha)+\frac{1}{2}\psi_{\ell}(x_{i};\alpha)^{2}\right]\\
 & =\frac{2}{n\sigma}\sum_{\ell=1}^{d}\sum_{i=1}^{n}\sum_{j=1}^{n}\alpha_{i}\exp\left(-\frac{\|x_{i}-x_{j}\|^{2}}{\sigma}\right)\left[-1+\frac{2}{\sigma}(x_{i\ell}-x_{j\ell})^{2}\right]\\
 & \qquad+\frac{2}{n\sigma^{2}}\sum_{\ell=1}^{d}\sum_{i=1}^{n}\left[\sum_{j=1}^{n}\alpha_{j}(x_{j\ell}-x_{i\ell})\exp\left(-\frac{\|x_{i}-x_{j}\|^{2}}{\sigma}\right)\right]^{2}.
\end{align*}
We now rewrite $J(\alpha)$ in matrix form. The expression for the
term $J(\alpha)$ being optimised is the sum of two terms. 

\textbf{First Term}:
\[
\sum_{\ell=1}^{d}\sum_{i=1}^{n}\sum_{j=1}^{n}\alpha_{i}\exp\left(-\frac{\|x_{i}-x_{j}\|^{2}}{\sigma}\right)\left[-1+\frac{2}{\sigma}(x_{i\ell}-x_{j\ell})^{2}\right]
\]
 We only need to compute
\begin{align*}
 & \sum_{i=1}^{n}\sum_{j=1}^{n}\alpha_{i}\exp\left(-\frac{\|x_{i}-x_{j}\|^{2}}{\sigma}\right)(x_{i\ell}-x_{j\ell})^{2}\\
= & \sum_{i=1}^{n}\sum_{j=1}^{n}\alpha_{i}\exp\left(-\frac{\|x_{i}-x_{j}\|^{2}}{\sigma}\right)\left(x_{i\ell}^{2}+x_{j\ell}^{2}-2x_{i\ell}x_{j\ell}\right).
\end{align*}
Define 
\[
x_{\ell}:=\left[\begin{array}{ccc}
x_{1\ell} & \hdots & x_{m\ell}\end{array}\right]^{\top}.
\]
The final term may be computed with the right ordering of operations,
\[
-2(\alpha\odot x_{\ell})^{\top}Kx_{\ell},
\]
where $\alpha\odot x_{\ell}$ is the entry-wise product. The remaining
terms are sums with constant row or column terms. Define $s_{\ell}:=x_{\ell}\odot x_{\ell}$
with components $s_{i\ell}=x_{i\ell}^{2}$. Then
\begin{align*}
\sum_{i=1}^{n}\sum_{j=1}^{n}\alpha_{i}k_{ij}s_{j\ell} & =\alpha^{\top}Ks_{\ell}.
\end{align*}
Likewise
\[
\sum_{i=1}^{n}\sum_{j=1}^{n}\alpha_{i}x_{i\ell}^{2}k_{ij}=(\alpha\odot s_{\ell})^{\top}K\mathbf{1}.
\]

\textbf{Second Term}: Considering only the $\ell$-th dimension, this
is
\begin{align*}
 & \sum_{i=1}^{n}\left[\sum_{j=1}^{n}\alpha_{j}(x_{j\ell}-x_{i\ell})\exp\left(-\frac{\|x_{i}-x_{j}\|^{2}}{\sigma}\right)\right]^{2}.
\end{align*}
In matrix notation, the inner sum is a column vector,
\[
K(\alpha\odot x_{\ell})-\left(K\alpha\right)\odot x_{\ell}.
\]
We take the entry-wise square and sum the resulting vector. Denote
by $D_{x}:=\text{diag}(x)$, then the following two relations hold

\begin{align*}
K(\alpha\odot x) & =KD_{x}\alpha,\\
(K\alpha)\odot x & =D_{x}K\alpha.
\end{align*}
This means that $J(\alpha)$ as defined previously,

\begin{align*}
J(\alpha) & =\frac{2}{n\sigma}\sum_{\ell=1}^{d}\left[\frac{2}{\sigma}\left[\alpha^{T}Ks_{\ell}+(\alpha\odot s_{\ell})^{\top}K\mathbf{1}-2(\alpha\odot x_{\ell})^{\top}Kx_{\ell}\right]-\alpha^{T}K\mathbf{1}\right]\\
 & +\frac{2}{n\sigma^{2}}\sum_{\ell=1}^{d}\left[(\alpha\odot x_{\ell})^{\top}K-x_{\ell}^{\top}\odot(\alpha^{\top}K)\right]\left[K(\alpha\odot x_{\ell})-(K\alpha)\odot x_{\ell}\right],
\end{align*}
can be rewritten as

\begin{align*}
J(\alpha) & =\frac{2}{n\sigma}\alpha^{T}\sum_{\ell=1}^{d}\left[\frac{2}{\sigma}(Ks_{\ell}+D_{s_{\ell}}K\mathbf{1}-2D_{x_{\ell}}Kx_{\ell})-K\mathbf{1}\right]\\
 & +\frac{2}{n\sigma^{2}}\alpha^{T}\left(\sum_{\ell=1}^{d}\left[D_{x_{\ell}}K-KD_{x_{\ell}}\right]\left[KD_{x_{\ell}}-D_{x_{\ell}}K\right]\right)\alpha\\
 & =\frac{2}{n\sigma}\alpha^{T}b+\frac{2}{n\sigma^{2}}\alpha^{\top}C\alpha,
\end{align*}
where

\begin{align*}
b & =\sum_{\ell=1}^{d}\left(\frac{2}{\sigma}(Ks_{\ell}+D_{s_{\ell}}K\mathbf{1}-2D_{x_{\ell}}Kx_{\ell})-K\mathbf{1}\right)\in\mathbb{R}^{n},\\
C & =\sum_{\ell=1}^{d}\left[D_{x_{\ell}}K-KD_{x_{\ell}}\right]\left[KD_{x_{\ell}}-D_{x_{\ell}}K\right]\in\mathbb{R}^{n\times n}.
\end{align*}
Assuming $C$ is invertible, this is minimised by \emph{
\[
\hat{\alpha}=-\frac{\sigma}{2}C^{-1}b.
\]
}
\end{proof}
As in \cite{SriFukKumGreHyv14}, we add a term $\lambda\Vert f\Vert_{{\cal H}}^{2}$
for $\lambda\in\mathbb{R}^{+}$, in order to control the norm of the
natural parameters in the RKHS $\Vert f\Vert_{{\cal H}}^{2}$. This
results in the regularised and numerically more stable solution $\hat{\alpha}_{\lambda}:=(C+\lambda I)^{-1}b$.

\subsection*{Reduced Computational Costs via Low-rank Approximations and Conjugate
Gradient}

Solving the linear system in \eqref{eq:lite_estimator} requires ${\cal O}(n^{3})$
computation and ${\cal O}(n^{2})$ storage for a fixed random sub-sample
of the chain history $\mathbf{z}.$ In order to allow for large $n$,
and to exploit potential manifold structure in the RKHS, we apply
a low-rank approximation to the kernel matrix via incomplete Cholesky
\cite[Alg. 5.12]{shawe2004kernel}, that is a standard way to achieve
linear computational costs for kernel methods. We rewrite the kernel
matrix 
\[
K\approx LL^{\top},
\]
where $L\in\mathbb{R}^{n\times\ell}$ is obtained via dual partial
Gram\textendash Schmidt orthonormalisation and costs both ${\cal O}(n\ell)$
computation and storage. Usually $\ell\ll n$, and $\ell$ can be
chosen via an accuracy cut-off parameter on the kernel spectrum in
the same fashion as for other low-rank approximations, such as PCA\footnote{In this paper, we solely use the Gaussian kernel, whose spectrum decays
exponentially fast.}. Given such a representation of $K$, we can rewrite any matrix-vector
product as 
\[
Kb\approx(LL^{\top})b=L(L^{\top}b),
\]
 where each left multiplication of $L$ costs ${\cal O}(n\ell)$ and
we never need to store $LL^{\top}$. This idea can be used to achieve
costs of ${\cal O}(n\ell)$ when computing $b$, and left-multiplying
$C$. Combining the technique with conjugate gradient (CG) allows
to solve \eqref{eq:lite_estimator} with a maximum of $n$ such matrix-vector
products, yielding a total computational cost of ${\cal O}(n^{2}\ell)$.
In practice, we can monitor residuals and stop CG after a fixed number
of iterations $\tau\ll n$, where $\tau$ depends on the decay of
the spectrum of $K$. We arrive at a total cost of ${\cal O}(n\ell\tau)$
computation and ${\cal O}(n\ell)$ storage. CG also has the advantage
of allowing for 'hot starts', i.e. initialising the linear solver
at a previous solution. Further details can be found in our implementation.

\section{Finite Feature Space Estimator}

\label{sub:Appendix_finite_details}

\subsubsection{Proof of Proposition \ref{prop:finite_estimator}}

We assume the model log-density \eqref{eq:score_matching_parametric_model}
takes the primal form in a finite dimensional feature space as in
Proposition \ref{prop:finite_estimator}, then again directly implement
score functions in \eqref{eq:score_matching_objective_sample_version}
and minimise it via a linear solve.
\begin{proof}
As assumed the log unnormalised density takes the form
\[
f(x)=\langle\theta,\phi_{x}\rangle_{{\cal H}_{m}}=\theta^{\top}\phi_{x},
\]
where $x\in\mathbb{R}^{d}$ is embedded into a finite dimensional
feature space ${\cal H}_{m}=\mathbb{R}^{m}$ as $x\mapsto\phi_{x}$.
The score functions in \eqref{eq:score_matching_objective_sample_version}
then can be written as the simple linear form
\begin{align}
\psi_{\ell}(\xi;\theta):=\frac{\partial\log\tilde{\pi}(x;\theta)}{\partial x_{\ell}} & =\theta^{\top}\dot{\phi}_{x}^{\ell}\quad\text{and}\quad\partial_{\ell}\psi_{\ell}(\xi;\theta):=\frac{\partial^{2}\log\tilde{\pi}(x;\theta)}{\partial x_{\ell}^{2}}=\theta^{\top}\ddot{\phi}_{x}^{\ell},\label{eq:score_function_fourier}
\end{align}
where we defined the $m$-dimensional feature vector derivatives $\dot{\phi}_{x}^{\ell}:=\frac{\partial}{\partial x_{\ell}}\phi_{x}$
and $\ddot{\phi}_{x}^{\ell}:=\frac{\partial^{2}}{\partial x_{\ell}^{2}}\phi_{x}$.
Plugging those into the empirical score matching objective in \eqref{eq:score_matching_objective_sample_version},
we arrive at
\begin{align}
J(\theta) & =\frac{1}{n}\sum_{i=1}^{n}\sum_{\ell=1}^{d}\left[\partial_{\ell}\psi_{\ell}(x_{i};\theta)+\frac{1}{2}\psi_{\ell}^{2}(x_{i};\theta)\right]\nonumber \\
 & =\frac{1}{n}\sum_{i=1}^{n}\sum_{\ell=1}^{d}\left[\theta^{T}\ddot{\phi}_{x_{i}}^{\ell}+\frac{1}{2}\theta^{T}\left(\dot{\phi}_{x_{i}}^{\ell}\left(\dot{\phi}_{x_{i}}^{\ell}\right)^{T}\right)\theta\right]\nonumber \\
 & =\frac{1}{2}\theta^{T}C\theta-\theta^{T}b\label{eq:score_match_objective_random_feats}
\end{align}
 where
\begin{equation}
b:=-\frac{1}{n}\sum_{i=1}^{n}\sum_{\ell=1}^{d}\ddot{\phi}_{x_{i}}^{\ell}\in\mathbb{R}^{m}\quad\text{and}\quad C:=\frac{1}{n}\sum_{i=1}^{n}\sum_{\ell=1}^{d}\left(\dot{\phi}_{x_{i}}^{\ell}\left(\dot{\phi}_{x_{i}}^{\ell}\right)^{\top}\right)\in\mathbb{R}^{m\times m}.\label{eq:b_and_C_random_feats}
\end{equation}
Assuming that $C$ is invertible (trivial for $n\geq m$), the objective
is uniquely minimised by differentiating \eqref{eq:score_match_objective_random_feats}
wrt. $\theta$, setting to zero, and solving for $\theta$. This gives
\begin{equation}
\hat{\theta}:=C^{-1}b.\label{eq:score_match_linear_system_random_feats}
\end{equation}

\end{proof}
Again, similar to \cite{SriFukKumGreHyv14}, we add a term $\lambda/2\Vert\theta\Vert_{2}^{2}$
for $\lambda\in\mathbb{R}^{+}$ to \eqref{eq:score_match_objective_random_feats},
in order to control the norm of the natural parameters $\theta\in{\cal H}^{m}$.
This results in the regularised and numerically more stable solution
$\hat{\theta}_{\lambda}:=(C+\lambda I)^{-1}b$.

Next, we give an example for the approximate feature space ${\cal H}_{m}$.
Note that the above approach can be combined with \emph{any} set of
finite dimensional approximate feature mappings $\phi_{x}$.

\subsection*{Example: Random Fourier Features for the Gaussian Kernel}

We now combine the finite dimensional approximate infinite dimensional
exponential family model with the ``random kitchen sink'' \cite{Rahimi2007}.
Assume a translation invariant kernel $k(x,y)=\tilde{k}(x-y)$. Bochner's
theorem gives the representation
\[
k(x,y)=\tilde{k}(x-y)=\int_{\mathbb{R}^{d}}\exp\left(i\omega^{\top}(x-y)\right)\mathrm{d}\Gamma(\omega),
\]
where $\Gamma(\omega)$ is the Fourier transform of the kernel. An
approximate feature mapping for such kernels can be obtained via dropping
imaginary terms and approximating the integral with Monte Carlo integration.
This gives 
\[
\phi_{x}=\sqrt{\frac{2}{m}}\left[\cos(\omega_{1}^{\top}x+u_{1}),\dots,\cos(\omega_{m}^{\top}x+u_{m})\right],
\]
with fixed random basis vector realisations that depend on the kernel
via $\Gamma(\omega)$,
\begin{align*}
\omega_{i}\sim\Gamma(\omega),
\end{align*}
and fixed random offset realisations 
\[
u_{i}\sim\texttt{Uniform}[0,2\pi],
\]
for $i=1\dots m$. It is easy to see that this approximation is consistent
for $m\to\infty$, i.e.
\[
\mathbb{E}_{\omega,b}\left[\phi_{x}^{T}\phi_{y}\right]=k(x,y).
\]
See \cite{Rahimi2007} for details and a uniform convergence bound
and \cite{sriperumbudurszabo15optimal} for a more detailed analysis
with tighter bounds. Note that it is possible to achieve logarithmic
computational costs in $d$ exploiting properties of Hadamard matrices
\cite{le2013fastfood}.

The feature map derivatives \eqref{eq:score_function_fourier} are
given by
\begin{align*}
\dot{\phi}_{\xi}^{\ell} & =\sqrt{\frac{2}{m}}\frac{\partial}{\partial\xi_{\ell}}\left[\cos(\omega_{1}^{T}\xi+u_{1}),\dots,\cos(\omega_{m}^{T}\xi+u_{m})\right]\\
 & =-\sqrt{\frac{2}{m}}\left[\sin(\omega_{1}^{T}\xi+u_{1})\omega_{1\ell},\dots,\sin(\omega_{m}^{T}\xi+u_{m})\omega_{m\ell}\right]\\
 & =-\sqrt{\frac{2}{m}}\left[\sin(\omega_{1}^{T}\xi+u_{1}),\dots,\sin(\omega_{m}^{T}\xi+u_{m})\right]\odot\left[\omega_{1\ell},\dots,\omega_{m\ell}\right],
\end{align*}
where $\omega_{j\ell}$ is the $\ell$-th component of $\omega_{j}$,
and
\begin{align*}
\ddot{\phi}_{\xi}^{\ell}: & =-\sqrt{\frac{2}{m}}\frac{\partial}{\partial\xi_{\ell}}\left[\sin(\omega_{1}^{T}\xi+u_{1}),\dots,\sin(\omega_{m}^{T}\xi+u_{m})\right]\odot\left[\omega_{1\ell},\dots,\omega_{m\ell}\right]\\
 & =-\sqrt{\frac{2}{m}}\left[\cos(\omega_{1}^{T}\xi+u_{1}),\dots,\cos(\omega_{m}^{T}\xi+u_{m})\right]\odot\left[\omega_{1\ell}^{2},\dots,\omega_{m\ell}^{2}\right]\\
 & =-\phi_{\xi}\odot\left[\omega_{1\ell}^{2},\dots,\omega_{m\ell}^{2}\right],
\end{align*}
where $\odot$ is the element-wise product. Consequently the gradient
is given by 
\[
\nabla_{\xi}\phi_{\xi}=\begin{bmatrix}\dot{\phi}_{\xi}^{1}\\
\vdots\\
\dot{\phi}_{\xi}^{d}
\end{bmatrix}\in\mathbb{R}^{d\times m}.
\]
As an example, the translation invariant Gaussian kernel and its Fourier
transform are
\[
k(x,y)={\cal \exp}\left(-\sigma^{-1}\Vert x-y\Vert_{2}^{2}\right)\quad\text{and}\quad\Gamma(\omega)={\cal N}\left(\omega\Big\vert\mathbf{0},\sigma^{-2}I_{m}\right).
\]

\subsection*{Constant Cost Updates}

\label{sub:rank_d_updates}

A convenient property of the finite feature space approximation is
that its primal representation of the solution allows to update \eqref{eq:b_and_C_random_feats}
in an online fashion. When combined with MCMC, each new point $x_{t+1}$
of the Markov chain history only adds a term of the form $-\sum_{\ell=1}^{d}\ddot{\phi}_{x_{t+1}}^{\ell}\in\mathbb{R}^{m}$
and $\sum_{\ell=1}^{d}\dot{\phi}_{x_{t+1}}^{\ell}(\dot{\phi}_{x_{t+1}}^{\ell})^{\top}\in\mathbb{R}^{m\times m}$
to the moving averages of $b$ and $C$ respectively. Consequently,
at iteration $t$, rather than fully re-computing \eqref{eq:score_match_linear_system_random_feats}
at the cost of ${\cal O}(tdm^{2}+m^{3})$ for every new point, we
can use rank-$d$ updates to construct the minimiser of \eqref{eq:score_match_objective_random_feats}
from the solution of the previous iteration. Assume we have computed
the sum of all moving average terms, 
\[
\bar{C}_{t}^{-1}:=\left(\sum_{i=1}^{t}\sum_{\ell=1}^{d}\left(\dot{\phi}_{x_{i}}^{\ell}\left(\dot{\phi}_{x_{i}}^{\ell}\right)^{\top}\right)\right)^{-1}
\]
from feature vectors derivatives $\ddot{\phi}_{x_{i}}^{\ell}\in\mathbb{R}^{m}$
of some set of points $\left\{ x_{i}\right\} _{i=1}^{t}$, and subsequently
receive receive a new point $x_{t+1}$. We can then write the inverse
of the new sum as
\begin{align*}
\bar{C}_{t+1}^{-1}: & =\left(\bar{C}_{t}+\sum_{\ell=1}^{d}\left(\dot{\phi}_{x_{t+1}}^{\ell}\left(\dot{\phi}_{x_{t+1}}^{\ell}\right)^{\top}\right)\right)^{-1}.
\end{align*}
This is the inverse of the rank-$d$ perturbed previous matrix $\bar{C}_{t}$.
We can therefore construct this inverse using $d$ successive applications
of the Sherman-Morrison-Woodbury formula for rank-one updates, each
using ${\cal O}(m^{2})$ computation. Since $\bar{C}_{t}$ is positive
definite\footnote{$C$ is the empirical covariance of the feature derivatives $\dot{\phi}_{x_{i}}^{\ell}$.},
we can represent its inverse as a numerically much more stable Cholesky
factorisation $\bar{C}_{t}=\bar{L}_{t}\bar{L}_{t}^{\top}$. It is
also possible to perform cheap rank-$d$ updates of such Cholesky
factors\footnote{We use the open-source implementation provided at \url{https://github.com/jcrudy/choldate}}.
Denote by $\bar{b}_{t}$ the sum of the moving average $b$. We solve
\eqref{eq:score_match_linear_system_random_feats} as 
\begin{align*}
\hat{\theta} & =C^{-1}b=\left(\frac{1}{t}\bar{C}_{t}\right)^{-1}\left(\frac{1}{t}\bar{b}_{t}\right)=\bar{C}_{t}^{-1}\bar{b}_{t}=\bar{L}_{t}^{-\top}\bar{L}_{t}^{-1}\bar{b}_{t},
\end{align*}
using cheap triangular back-substitution from $\bar{L}_{t}$, and
never storing $\bar{C}_{t}^{-1}$ or $\bar{L}_{t}^{-1}$ explicitly.

Using such updates, the computational costs for updating the approximate
infinite dimensional exponential family model in \emph{every }iteration
of the Markov chain are ${\cal O}(dm^{2})$, which \emph{constant
in $t$. }We can therefore use \emph{all} points in the history for
constructing a proposal. See our implementation for further details.

\subsubsection{Algorithmic Description:}
\begin{enumerate}
\item Update sums 
\[
\bar{b}_{t+1}=\bar{b}_{t}-\sum_{\ell=1}^{d}\ddot{\phi}_{x_{t+1}}^{\ell}\quad\text{and}\quad\bar{C}_{t+1}=\bar{C}_{t}+\frac{1}{2}\sum_{\ell=1}^{d}\dot{\phi}_{x_{t+1}}^{\ell}(\dot{\phi}_{x_{t+1}}^{\ell})^{\top}.
\]
 
\item Perform rank-$d$ update to obtain updated Cholesky factorisation
$\bar{L}_{t+1}\bar{L}_{t+1}^{T}=\bar{C}_{t+1}$.
\item Update approximate infinite dimensional exponential family parameters
\end{enumerate}
\[
\hat{\theta}=\bar{L}_{t+1}^{-\top}\bar{L}_{t+1}^{-1}\bar{b}_{t+1}.
\]

\section{Ergodicity of KMC lite}

\label{sub:Appenidx_ergodicity_lite_proof}

\paragraph{Notation}

Denote by $\alpha(x_{t},x^{*}(p'))$ the probability of accepting
a $(p',x^{*})$ proposal at state $x_{t}$. Let $a\wedge b=\min(a,b)$.
Define $c(x^{(0)}):=L\epsilon^{2}\nabla\log\pi(x^{(0)})/2+\epsilon^{2}\sum_{i=1}^{L-1}(L-i)\nabla\log\pi(x^{(i\epsilon)})$
and $d(x^{(0)}):=\epsilon(\nabla f(x^{(0)})+\nabla f(x^{(L\epsilon)}))/2+\epsilon\sum_{i=1}^{L-1}\nabla f(x^{(i\epsilon)})$,
where $x^{(i\epsilon)}$ is the $i$-th point of the leapfrog integration
from $x=x^{(0)}$.

\subsubsection{Proof of Proposition \ref{prop:ergodicity_kmc_lite}}
\begin{proof}
We assumed $\pi(x)$ is log-concave in the tails, meaning $\exists x_{U}>0$
s.t. for $x^{*}>x_{t}>x_{U}$, we have $\pi(x^{*})/\pi(x_{t})\leq e^{-\alpha_{1}(\Vert x^{*}\Vert_{2}-\Vert x_{t}\Vert_{2})}$
and for $x_{t}>x^{*}>x_{U}$, we have $\pi(x^{*})/\pi(x_{t})\geq e^{-\alpha_{1}(\Vert x^{*}\Vert_{2}-\Vert x_{t}\Vert_{2})}$,
and a similar condition holds in the negative tail. Furthermore, we
assumed fixed HMC parameters: $L$ leapfrog steps of size $\epsilon$,
and wlog the identity mass matrix $I$. Following \cite{roberts1996geometric,mengersen1996rates},
it is sufficient to show 
\[
\limsup_{\Vert x_{t}\Vert_{2}\to\infty}\int\left[e^{s(\Vert x^{*}(p')\Vert_{2}-\Vert x_{t}\Vert_{2})}-1\right]\alpha(x_{t},x^{*}(p'))\mu(dp')<0,
\]
for some $s>0$, where $\mu(\cdot)$ is a standard Gaussian measure.
Denoting the integral $I_{-\infty}^{\infty}$, we split it into 
\[
I_{-\infty}^{-x_{t}^{\delta}}+I_{-x_{t}^{\delta}}^{x_{t}^{\delta}}+I_{x_{t}^{\delta}}^{\infty},
\]
for some $\delta\in(0,1)$. We show that the first and third terms
decay to zero whilst the second remains strictly negative as $x_{t}\to\infty$
(a similar argument holds as $x_{t}\to-\infty$). We detail the case
$\nabla f(x)\uparrow0$ as $x\to\infty$ here, the other is analogous.
Taking $I_{-x_{t}^{\delta}}^{x_{t}^{\delta}}$, we can choose an $x_{t}$
large enough that $x_{t}-C-L\epsilon x_{t}^{\delta}>x_{U}$, $-\gamma_{1}<c(x_{t}-x_{t}^{\delta})<0$
and $-\gamma_{2}<d(x_{t}-x_{t}^{\delta})<0$. So for $p'\in(0,x_{t}^{\delta})$
we have 
\[
L\epsilon p'>x^{*}-x_{t}>L\epsilon p'-\gamma_{1}\implies e^{-\alpha_{1}(-\gamma_{1}+L\epsilon p')}\geq e^{-\alpha_{1}(x^{*}-x_{t})}\geq\pi(x^{*})/\pi(x_{t}),
\]
where the last inequality comes from the log-concave tails assumption.
For $p'\in(\gamma_{2}^{2}/2,x_{t}^{\delta})$ 
\[
\alpha(x_{t},x^{*})\leq1\wedge\frac{\pi(x^{*})}{\pi(x_{t})}\exp\left(p'\gamma_{2}/2-\gamma_{2}^{2}/2\right)\leq1\wedge\exp\left(-\alpha_{2}p'+\alpha_{1}\gamma_{1}-\gamma_{2}^{2}/2\right),
\]
where $x_{t}$ is large enough that $\alpha_{2}=\alpha_{1}L\epsilon-\gamma_{2}/2>0$.
Similarly for $p'\in(\gamma_{1}/L\epsilon,x_{t}^{\delta})$ 
\[
e^{sL\epsilon p'}-1\geq e^{s(x^{*}-x_{t})}-1\geq e^{s(L\epsilon p'-\gamma_{1})}-1>0.
\]
Because $\gamma_{1}$ and $\gamma_{2}$ can be chosen to be arbitrarily
small, then for large enough $x_{t}$ we will have 
\begin{align}
0<I_{0}^{x_{t}^{\delta}} & \leq\int_{\gamma_{1}/L\epsilon}^{x_{t}^{\delta}}[e^{sL\epsilon p'}-1]\exp\left(-\alpha_{2}p'+\alpha_{1}\gamma_{1}-\gamma_{2}^{2}/2\right)\mu(dp')+I_{0}^{\gamma_{1}/L\epsilon}\nonumber \\
 & =e^{c_{1}}\int_{\gamma_{1}/L\epsilon}^{x_{t}^{\delta}}[e^{s_{2}p'}-1]e^{-\alpha_{2}p'}\mu(dp')+I_{0}^{\gamma_{1}/L\epsilon},\label{eqn1}
\end{align}
where $c_{1}=\alpha_{1}\gamma_{1}-\gamma_{2}^{2}/2>0$ for large enough
$x_{t}$, as $\gamma_{1}$ and $\gamma_{2}$ are of the same order.
Now turning to $p'\in(-x_{t}^{\delta},0)$, we can use an exact rearrangement
of the same argument (noting that $c_{1}$ can be made arbitrarily
small) to get 
\begin{equation}
I_{-x_{t}^{\delta}}^{0}\leq e^{c_{1}}\int_{\gamma_{1}/L\epsilon}^{x_{t}^{\delta}}[e^{-s_{2}p'}-1]\mu(dp')<0.\label{eqn2}
\end{equation}
Combining \eqref{eqn1} and \eqref{eqn2} and rearranging as in \cite[Theorem 3.2]{mengersen1996rates}
shows that $I_{-x_{t}^{\delta}}^{x_{t}^{\delta}}$ is strictly negative
in the limit if $s_{2}=sL\epsilon$ is chosen small enough, as $I_{0}^{\gamma_{2}/L\epsilon}$
can also be made arbitrarily small.

For $I_{-\infty}^{-x_{t}^{\delta}}$ it suffices to note that the
Gaussian tails of $\mu(\cdot)$ will dominate the exponential growth
of $e^{s(\Vert x^{*}(p')\Vert_{2}-\Vert x_{t}\Vert_{2})}$ meaning
the integral can be made arbitrarily small by choosing large enough
$x_{t}$, and the same argument holds for $I_{x_{t}^{\delta}}^{\infty}$.
\end{proof}

\section{Additional Experimental Details}

\label{sec:app_experimental_details}This section contains additional
details for the experiments in Section \ref{sec:Experiments}.

\subsection{Stability in High Dimensions}

\label{sub:app_stability_in_high_dim}

\begin{figure}
\begin{centering}
\includegraphics[clip,scale=0.5]{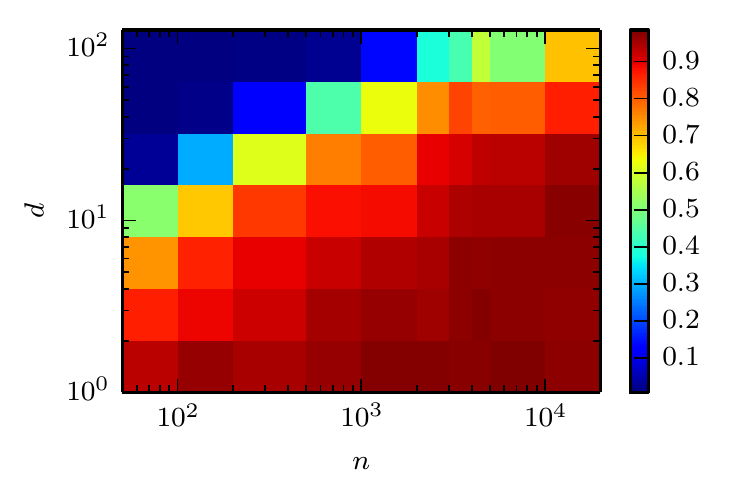}\includegraphics[scale=0.5]{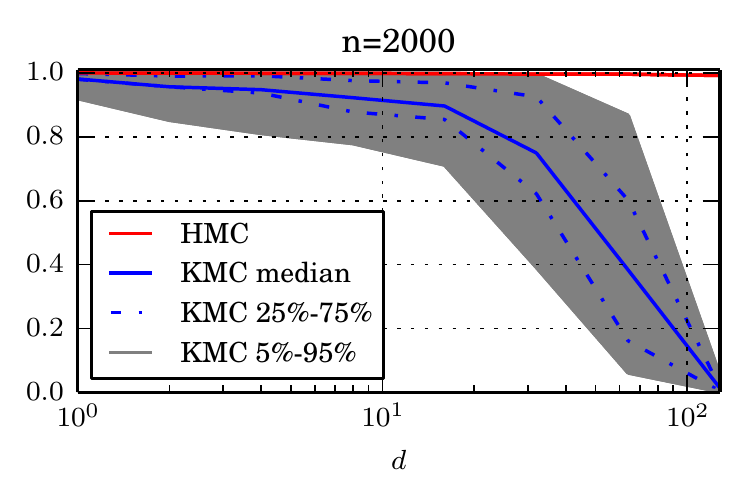}\includegraphics[bb=0bp 0bp 216bp 144bp,scale=0.5]{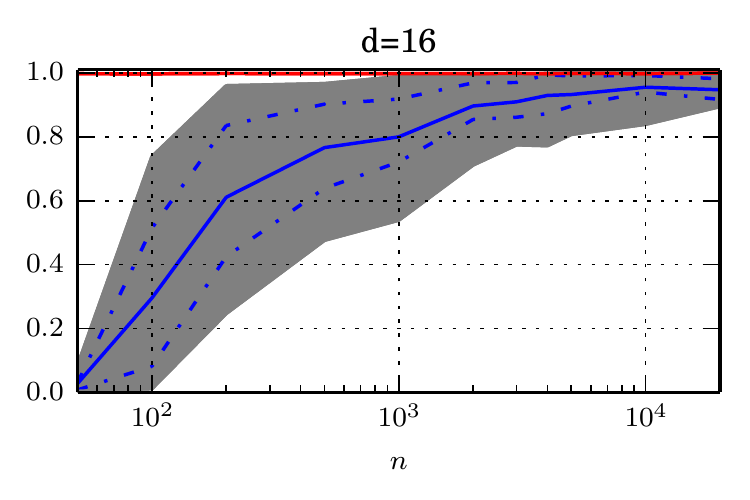}
\par\end{centering}

\caption{\label{fig:kmc_trajectories_mean_acceptance_isotropic}Acceptance
probability of kernel induced Hamiltonian flow for a standard Gasussian
in high dimensions for an isotropic Gaussian. \textbf{Left:} As a
function of\textbf{ $n=m$} (x-axis) and $d$ (y-axis). \textbf{Middle:
}Slices through left plot with error bars for a fixed $n=m$ and as
a function in \textbf{$d$} (left), and for a fixed $d$ as a function
of $n=m$ (right). }
\end{figure}

We reproduce the experiment in Figure \ref{fig:kmc_trajectories_mean_acceptance}
on an \emph{isotropic} Gaussian in increasing dimension. As length-scales
across all principal components are equal, this is a significantly
less challenging target to estimate gradients for; though still useful
as a benchmark representing very smooth targets. We use a standard
Gaussian kernel and the same experimental protocol as for Figure \ref{fig:kmc_trajectories_mean_acceptance}.
The estimator works slightly better than on the target considered
in Figure \ref{fig:kmc_trajectories_mean_acceptance}, and performs
well up to $d\approx100$, see Figure \ref{fig:kmc_trajectories_mean_acceptance_isotropic}.

\subsection{Banana target}

\label{sub:app_banana_target}

Following \cite{sejdinovic_kernel_2014,Haario1999}, let $X\sim\mathcal{N}(0,\Sigma)$
be a multivariate normal in $d\geq2$ dimensions, with $\Sigma=\text{{diag}}(v,1,\ldots,1)$,
which undergoes the transformation $X\to Y$, where $Y_{2}=X_{2}+b(X_{1}^{2}-v)$,
and $Y_{i}=X_{i}$ for $i\neq2$. We will write $Y\sim\mathcal{B}(b,v)$.
It is clear that $\mathbb{E}Y=0$, and that 

\[
\mathcal{B}(y;b,v)=\mathcal{N}(y_{1};0,v)\mathcal{N}(y_{2};b(y_{1}^{2}-v),1)\prod_{j=3}^{d}\mathcal{N}(y_{j};0,1).
\]
We choose $d=8$, $V=100$ and $b=0.03$, which corresponds to the
`strongly twisted' 8-dimensional Banana in \cite{sejdinovic_kernel_2014,Haario1999}.
The target is challenging due to the nonlinear dependence of the first
two dimensions and the highly position dependent scaling within these
dimensions.

\subsection{Pseudo-Marginal MCMC for GP Classification}

\label{sub:app_gp_target}

\paragraph{Model}

Closely following \cite{sejdinovic_kernel_2014}, we consider a joint
distribution of GP-latent variables $\mathbf{f}$, labels $\mathbf{y}$
(with covariate matrix \textbf{$X$}), and hyper-parameters $\theta$,
given by
\[
p(\mathbf{f},\mathbf{y},\theta)=p(\theta)p(\mathbf{f}|\theta)p(\mathbf{y}|\mathbf{f}),
\]
where $\mathbf{f}|\theta\sim{\cal N}(0,\mathcal{K}_{\theta})$, with
$\mathcal{K}_{\theta}$ modeling the covariance between latent variables
evaluated at the input covariates: $(\mathcal{K}_{\theta})_{ij}=\kappa(\mathbf{x}_{i},\mathbf{x}_{j}'|\theta)=\exp\left(-\frac{1}{2}\sum_{d=1}^{D}\frac{(x_{i,d}-x'_{j,d})^{2}}{\ell_{d}^{2}}\right)$
and $\theta_{d}=\log\ell_{d}^{2}$. This covariance parametrisation
allows to perform Automatic Relevance Determination. We here restrict
our attention to the binary logistic classifier, i.e. the likelihood
is given by 
\[
p(y_{i}|f_{i})=\frac{1}{1-\exp(-y_{i}f_{i})},
\]
where $y_{i}\in\{-1,1\}$. Aiming for a fully Bayesian treatment,
we wish to estimate the \emph{marginal }posterior of the hyper-parameters
$\theta$, motivated in \cite{FilipponeIEEETPAMI13}. The marginal
likelihood $p(\mathbf{y}|\theta)$ is intractable for non-Gaussian
likelihoods $p(\mathbf{y}|\mathbf{f})$, but can be replaced with
an unbiased estimate

\begin{equation}
\hat{p}(\mathbf{y}|\theta):=\frac{1}{n_{\textrm{imp}}}\sum_{i=1}^{n_{\textrm{imp}}}p(\mathbf{y}|\mathbf{f}^{(i)})\frac{p(\mathbf{f}^{(i)}|\theta)}{q(\mathbf{f}^{(i)}|\theta)},
\end{equation}
where $\left\{ \mathbf{f}^{(i)}\right\} _{i=1}^{n_{\textrm{imp}}}\sim q(\mathbf{f}|\theta)$
are $n_{imp}$ importance samples. In \cite{FilipponeIEEETPAMI13},
the importance distribution $q(\mathbf{f}|\theta)$ is chosen as the
Laplacian or as the Expectation Propagation (EP) approximation of
$p(\mathbf{f}|\mathbf{y},\theta)\propto p(\mathbf{y}|\mathbf{f})p(\mathbf{f}|\theta)$,
leading to state-of-the-art results.

\paragraph{Experimental details}

We here use a Laplace approximation and $n_{\text{imp}}=100$. We
consider classification of window against non-window glass in the
UCI Glass dataset, which induces a posterior that has a nonlinear
shape \cite[Figure 3]{sejdinovic_kernel_2014}. Since the ground truth
for the hyperparameter posterior is not available, we initially run
multiple hand-tuned standard Metropolis-Hastings chains for 500,000
iterations (with a 100,000 burn-in), keep every 1000-th sample in
each of the chains, and combine them. The resulting samples are used
as a benchmark, to evaluate the performance all algorithms. We use
the MMD between each sampler output and the benchmark sample is computed,
using the polynomial kernel $\left(1+\left\langle \theta,\theta'\right\rangle \right)^{3}$.
This corresponds to the estimation error of all mixed moments of order
up to 3.

\paragraph{Cross-validation}

Kernel parameters are tuned using a black box Bayesian optimisation
package\footnote{We use the open-source package \texttt{pybo}, available under \url{https://github.com/mwhoffman/pybo}}
and the median heuristic for KMC and KAMH repsectively. The Bayesian
optimisation uses standard parameters and is stopped after 15 iterations,
where each trial is done via a 5-fold cross-validation of the score
matching objective \eqref{eq:score_matching_objective_sample_version}.
We learn parameters after MCMC 500 iterations, and then re-learn after
2000. We tried re-learning parameters after more iterations, but this
did not lead to significant changes. The costs for this are neglectable
in the context of PM-MCMC as estimating the marginal likelihood takes
significantly more time than generating the KMC proposal.

\subsection{ABC MCMC}

\label{sub:app_abc_mcmc}In this section, we give a brief background
on Approximate Bayesian Computation, and how KMC can be used within
the framework. We then give details of the competing approach in the
final experiment in Section \ref{sub:experiment_abc}, including experimental
details and an analytic counterexample.

\paragraph{Likelihood-free Models}

Approximate Bayesian Computation is a method for inference in the
scenario where conditional on some parameter of interest $\theta$,
we can easily simulate data $x\sim f(\cdot|\theta)$, but for which
the likelihood function $f$ is unavailable \cite{Sisson2010}. We
however have data $y$ which assume to be from the model, and we have
a prior $\pi_{0}(\theta)$. A simple ABC algorithm is to sample $\theta_{i}\sim\pi_{0}(\cdot)$
(or any other suitable distribution), simulate data $x_{i}\sim f(\cdot|\theta_{i})$,
and `accept' $x_{i}$ as a sample from the approximate posterior $\pi_{\epsilon}(\theta|y)$
if $d(y,x)\leq\epsilon$. This procedure can be formalised by defining
the approximate likelihood as 
\begin{equation}
f_{\epsilon}(y|\theta)\propto\int g_{\epsilon}(y|x,\theta)f(x|\theta)\mathrm{d}x,\label{eqn:lik}
\end{equation}
where $g_{\epsilon}(y|x,\theta)$ is an appropriate kernel that gives
more importance to points for which $d(y,x)$ is smaller. In the simple
case above $g_{\epsilon}(y|x,\theta)=\mathbf{1}_{\{d(y,x)\leq\epsilon\}}$.
The ABC posterior is then found using $\pi_{\epsilon}(\theta|y)\propto f_{\epsilon}(y|\theta)\pi_{0}(\theta)$.
Often $g_{\epsilon}$ is based on some low-dimensional summary statistics,
which can have both advantages and disadvantages.

\paragraph{Likelihood-free MCMC}

There are many different way to do ABC, and clearly not all involve
Markov chain Monte Carlo. If the posterior however is not similar
to the prior, and if $\theta$ is more than three or four dimensional,
MCMC is a sensible option. Since the likelihood \eqref{eqn:lik} is
intractable, typically algorithms are considered for which an approximation
to either the likelihood, or the ABC posterior are used either in
constructing proposals, defining Metropolis-Hastings acceptance rates,
or both. We focus here on samplers which target $\pi_{\epsilon}(\theta|y)$
directly, c.f. \cite{marjoram2003markov}.

\paragraph{Pseudo-Marginal Metropolis-Hastings}

Similar to the approach taken in Section \ref{sub:app_gp_target},
we here accept proposals $\theta'\sim Q(\theta,\cdot)$ where $Q$
is some proposal mechanism (i.e. KMC), via replacing the likelihood
with an unbiased estimate. We accept according to the ratio 
\begin{equation}
\tilde{\alpha}(\theta,\theta')=\frac{\tilde{\pi}_{\epsilon}(\theta'|y)Q(\theta|\theta')}{\tilde{\pi}_{\epsilon}(\theta|y)Q(\theta'|\theta)},\label{eqn:ratio}
\end{equation}
where $\tilde{\pi}_{\epsilon}(\theta|y)=\pi_{0}(\theta)\tilde{g}_{\epsilon}(y|\theta),$
and 
\[
\tilde{g}_{\epsilon}(y|\theta)=\frac{1}{n_{\text{lik}}}\sum_{i}g_{\epsilon}(y|x_{i},\theta),~~\{x_{i}\}_{i=1}^{n_{\text{lik}}}\sim f(\cdot|\theta_{i})
\]
is a simple Monte Carlo estimator for the intractable likelihood \eqref{eqn:lik}.
Since it is easy to simulate from $f$ then $\tilde{g}_{\epsilon}(y|\theta)$
is typically easy to compute. As with other general Pseudo-Marginal
schemes, and as mentioned below the KMC acceptance \eqref{eq:kmc_accept_prob},
it is crucial that if $\theta'$ is accepted, the same estimate for
$\tilde{\pi}(\theta'|y)$ is used on the denominator of the Hastings
ratio in future iterations until the next proposal is accepted for
the scheme, c.f. \cite[Table 1]{Andrieu2009a}.

We can directly adapt KMC to the ABC case via plugging in the estimated
likelihood $\tilde{g}_{\epsilon}$ in the KMC acceptance ratio \eqref{eq:kmc_accept_prob}.

\paragraph{Synthetic Likelihood Metropolis-Hastings}

Following \cite{Wood:2010aa}, one idea to approximate the intractable
likelihood is to draw $n_{\text{lik}}$ samples $x_{i}\sim f(\cdot|\theta_{i})$,
and fit a Gaussian approximation to $f$, producing estimates $\hat{\mu}$
and $\hat{\Sigma}$ for the mean and covariance using $\{x_{i}\}_{i=1}^{n_{\text{lik}}}$.
If the error functon $g_{\epsilon}$ is also chosen to be a Gaussian
(with mean $y$ and variance $\epsilon$), then the marginal likelihood
$f_{\epsilon}(y|\theta)$ can be approximated as 
\[
y|\theta\sim\mathcal{N}\left(\hat{\mu},\hat{\Sigma}+\epsilon^{2}I\right)
\]
The likelihood is essentially approximated by a Gaussian $f_{G}$,
producing a synthetic posterior $\pi_{s}(\cdot)$, which is then used
in the accept-reject step. Clearly some approximation error is introduced
by the Gaussian likelihood approximation step, but as shown in \cite{Wood:2010aa},
it can be a reasonable choice for some models.

\paragraph{Hamiltonian ABC}

Introduced in \cite{Meeds2015}, the synthetic likelihood formulation
is used to construct a proposal, with the accept-reject step removed
altogether. Hamiltonian dynamics use the gradient $\nabla\log\pi(\theta)$
to suggest candidate values for the next state of a Markov chain which
are far from the current point, thus increasing the chances that the
chain mixes quickly. Here the gradient of the log-likelihood is unavailable,
so is approximated with that of a Gaussian (since the map $\theta\to(\mu,\Sigma)$
is not always clear this is done numerically, using a stochastic finite
differences estimate of the gradient, SPAS \cite[Sec. 4.3, 4.4]{Meeds2015}),
giving 
\[
\nabla\log\pi(\theta)\approx\sum_{i=1}^{n_{\text{lik}}}\nabla\log f_{G}(y_{i}|\hat{\mu},\hat{\Sigma})+\nabla\log\pi_{0}(\theta).
\]
Since there is no accept-reject step, the synthetic posterior is also
the target of this scheme (although there is also further bias introduced
by discretisation error), but the introduction of gradient-based dynamics
is hoped to improve mixing and hence efficiency of inferences compared
to random-walk type schemes.

\paragraph{A Counter-example}

\begin{figure}
\begin{centering}
\includegraphics[bb=0bp 0bp 504bp 504bp,clip,scale=0.34]{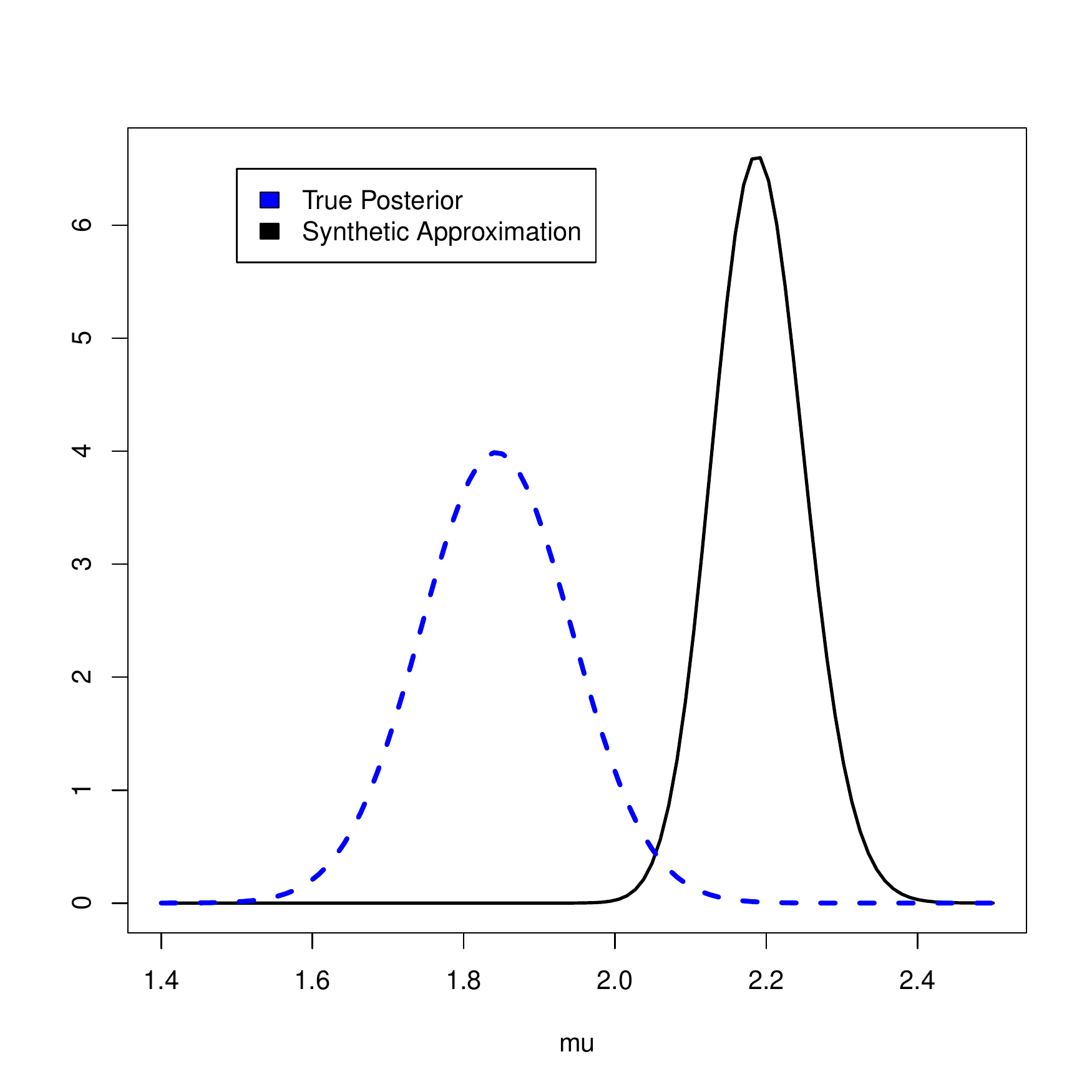} 
\par\end{centering}

\caption{\label{fig:experiment_gp_abc_counterexample}Counter example showing
posterior and its synthetic approximation for a simple toy model.}
\end{figure}

We give a very simple toy model to highlight the bias introduced by
the Hamiltonian ABC sampler. Consider posterior inference for the
mean parameter in a log-Normal model. Specifically, the true model
is 
\begin{align*}
\mu & \sim\mathcal{N}(\mu_{0},\tau_{0}),\\
y|\mu,\tau & \sim\log\mathcal{N}(\mu,\tau),
\end{align*}
where the precision $\tau$ and hyper-parameters $\mu_{0},\tau_{0}$
are known. The model is in fact conjugate, giving a Gaussian posterior
\[
\mu|y\sim\mathcal{N}\left(\frac{\tau_{0}\mu_{0}+\tau\sum_{i}\log x_{i}}{\tau_{0}+n\tau},\tau_{0}+n\tau\right).
\]
If we introduce a Gaussian approximation to the likelihood, then the
mean and precision of this approximation $f_{G}$ are (empirical estimates
for) 
\[
\mu_{G}=e^{\mu+1/2\tau},~~\tau_{G}=1/\text{Var}[Y_{i}]=\frac{e^{-2\mu-1/\tau}}{e^{1/\tau}-1},
\]
which depend on the current value for $\mu$ in the chain. The resulting
synthetic posterior is no longer tractable, but since it is one dimensional
we can approximate it numerically. Using $\mu_{0}=0$, $\tau_{0}=1/100$,
$\epsilon=0.1$ and $\tau=1$ then the true and approximate posteriors
for 100 data points generated using the truth $\mu=2$ are shown in
Figure \ref{fig:experiment_gp_abc_counterexample}. This is a proof
of concept that a likelihood with a positive skew being approximated
by a Gaussian introduces an upwards bias to the posterior.

\paragraph{Experimental details}

The simulation study in Section \ref{sub:experiment_abc} uses a slightly
more complex and multi-dimensional simulation example: a 10-dimensional
multivariate skew-Normal distribution, given by 
\[
p(y|\theta)=2\mathcal{N}\left(\theta,I\right)\Phi\left(\left\langle \alpha,y\right\rangle \right)
\]
with $\theta=\alpha=\mathbf{1}\cdot10$. In each iteration of KMC,
the likelihood is estimated via simulating $n_{\text{lik}}=10$ samples
from the above likelihood. We use the mean of all samples as summary
statistic, and a Gaussian similarity kernel $g_{\epsilon}(y|x,\theta)$
with a fixed $\epsilon=0.55$. Both KMC and HABC use a standard Gaussian
momentum, a uniformly random stepsize in $[0.01,0.1]$ and $L=50$
leapfrog steps. HABC is used with the suggested `sticky random numbers'
\cite[Section 4.4]{Meeds2015}, i.e. we use the same seed for all
simulations along a single proposal trajectory. Both algorithms are
run for $200+5000$ MCMC iterations. KMC then attempts to re-learn
smoothness parameters, and stops adaptation. Burn-in samples are discarded
when quantifying performance of all algorithms.

\paragraph{Friction, mixing, and number of simulations}

\begin{figure}
\begin{centering}
\includegraphics[scale=0.85]{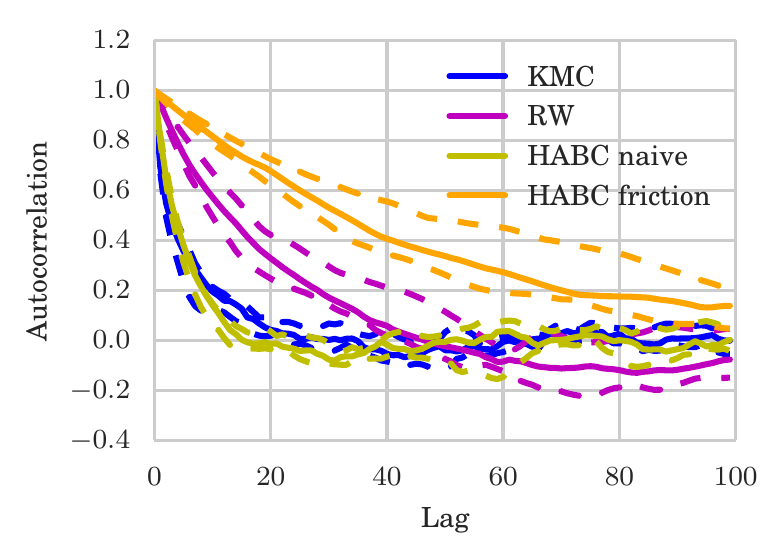}\includegraphics[scale=0.85]{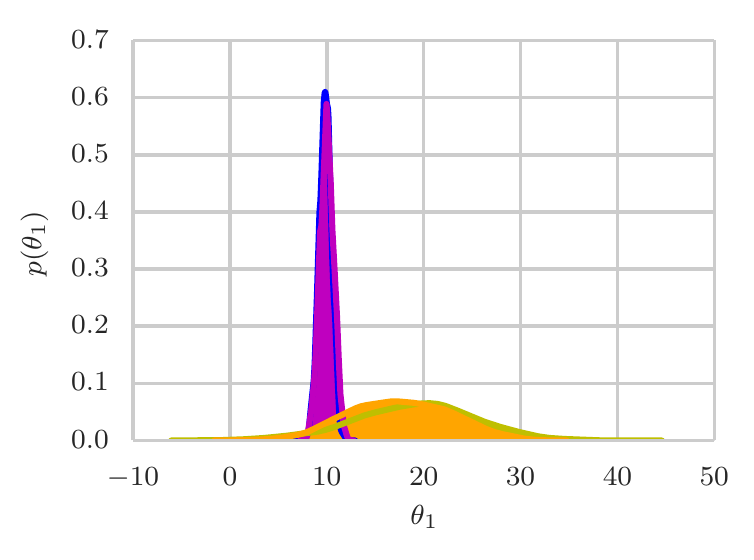}
\par\end{centering}

\caption{\label{fig:experiment_gp_abc_with_friction}\textbf{Left: }Counterexample
showing posterior and its synthetic approximation for a simple toy
model \textbf{Middle/right:} The same results as in Figure \ref{fig:experiment_gp_abc},
i.e. autocorrelation and marginal posterior for $\theta_{1}$, but
here we also show performance of HABC with added friction, which has
a severely negative impact on mixing. }
\end{figure}

HABC is used in its `stochastic gradient' \cite{ChenFoxGuestrin2014}
and has a `friction' parameter that we estimate using a running average
of the global covariance of all SPAS gradient evaluations, \cite[Equation 21]{Meeds2015}.
Note that we ran HABC with both the friction term included and removed,
where we found that adding friction has severely negative impact on
mixing, where not adding friction results in a wider posterior (with
the same bias). Figure \ref{fig:experiment_gp_abc} (middle, right)
show the results \emph{without} friction, Figure \ref{fig:experiment_gp_abc_with_friction}
shows the same plots with friction. We refer to our implementation
for further details. 

Due to the gradient estimation in every of the $L=50$ leap-frog steps,
e\emph{very} MCMC proposal for HABC requires $2L=100$ simulations
to be generated. In contrast, KMC only requires a single simulation,
for evaluating the accept/reject probability \eqref{eq:kmc_accept_prob}.
We leave studying the exact trade-offs of KMC's learning phase and
its ability to mix well as compared to HABC to future work.


\end{document}